\documentclass{article}

\usepackage{microtype}
\usepackage{multirow}
\usepackage{graphicx}
\usepackage{subfigure}
\usepackage[table,dvipsnames]{xcolor}
\usepackage{nicematrix}
\usepackage{enumitem}
\usepackage{soul}
\usepackage{booktabs} 
\usepackage{subcaption}

\usepackage{hyperref}

\usepackage[accepted]{icml2025}

\usepackage{amsmath}
\usepackage{amssymb}
\usepackage{mathtools}
\usepackage{amsthm}
\usepackage{xspace}

\newcommand{\name}{POPri\xspace}
\newcommand{\benchname}{LargeFedBench\xspace}

\usepackage[capitalize,noabbrev]{cleveref}

\theoremstyle{plain}

\theoremstyle{definition}

\theoremstyle{remark}

\definecolor{bettergreen}{RGB}{34,139,34} 
\definecolor{betterred}{RGB}{200,0,0}
\usepackage[textsize=tiny]{todonotes}

\icmltitlerunning{POPri: Private Federated Learning using Preference-Optimized Synthetic Data}

\begin{document}

\twocolumn[
\icmltitle{POPri: Private Federated Learning using Preference-Optimized Synthetic Data}



\icmlsetsymbol{equal}{*}


\begin{icmlauthorlist}
\icmlauthor{Charlie Hou}{equal,cmu}
\icmlauthor{Mei-Yu Wang}{equal,psc}
\icmlauthor{Yige Zhu}{equal}
\icmlauthor{Daniel Lazar}{cold}
\icmlauthor{Giulia Fanti}{cmu}
\end{icmlauthorlist}
\icmlaffiliation{psc}{Pittsburgh Supercomputing Center, Pittsburgh, USA}
\icmlaffiliation{cmu}{Department of ECE, Carnegie Mellon University, Pittsburgh, PA}
\icmlaffiliation{cold}{Coldrays, Tucson, AZ}

\icmlcorrespondingauthor{Charlie Hou}{hou.charlie2@gmail.com}



\icmlkeywords{differential privacy, synthetic data, federated learning, preference optimization}

\vskip 0.3in
]



\printAffiliationsAndNotice{\icmlEqualContribution} 

\begin{abstract}
In practical settings, differentially private federated learning (DP-FL) is the dominant method for training models from private, on-device client data. 
Recent work has suggested that DP-FL may be enhanced or outperformed by methods that use DP synthetic data \cite{wu2024prompt,PrE-Text}.
The primary algorithms for generating DP synthetic data for FL applications require careful prompt engineering based on public information and/or iterative private client feedback. 
Our key insight is that the private client feedback collected by prior DP synthetic data methods \cite{PrE-Text,xie2024differentially} can be viewed as an RL (reinforcement learning) reward. 
Our algorithm, Policy
Optimization for Private Data (POPri) harnesses client feedback using policy optimization algorithms such as Direct Preference Optimization (DPO) 
to fine-tune LLMs to generate high-quality DP synthetic data. 
To evaluate \name, we release \benchname, a new federated text benchmark for uncontaminated LLM evaluations on federated client data. 
\name closes the gap in performance between the fully-private and non-private settings by up to 58$\%$, compared to 28$\%$ for prior synthetic data methods, and 3$\%$ for state-of-the-art DP federated learning methods. The code and data are available at \url{https://github.com/meiyuw/POPri}.  
\end{abstract}

\begin{figure*}[t]
\centering
\hbox{
\includegraphics[width=11cm]{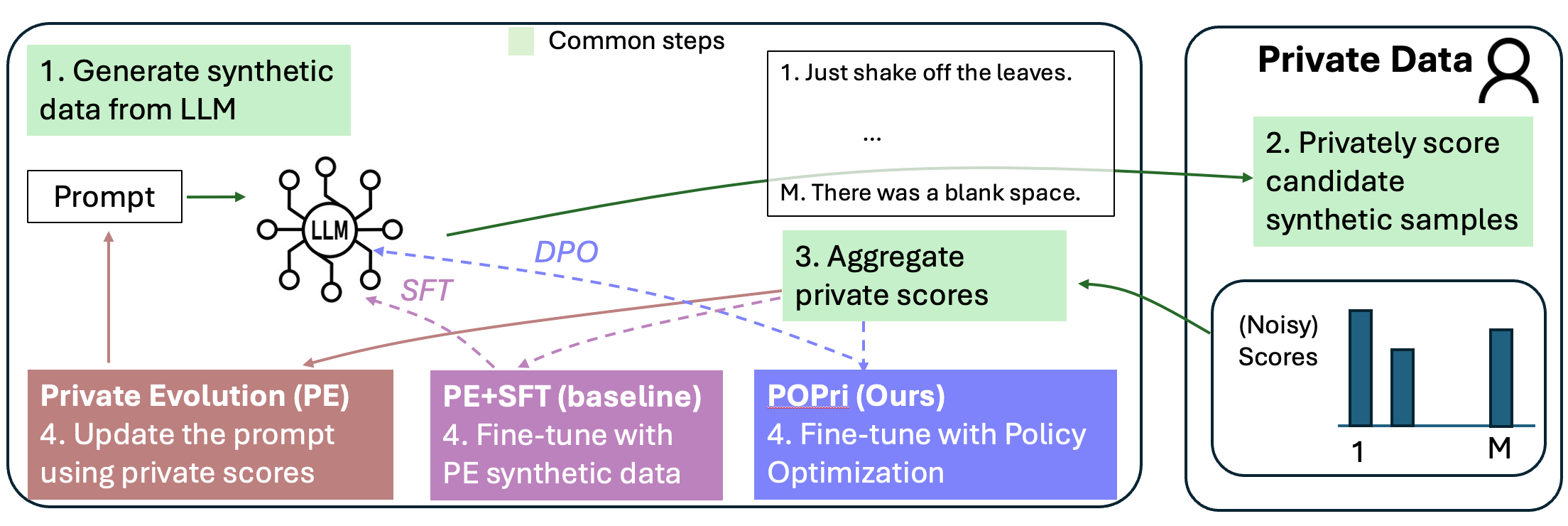}
\hskip 0.1in
\includegraphics[width=5.75cm]{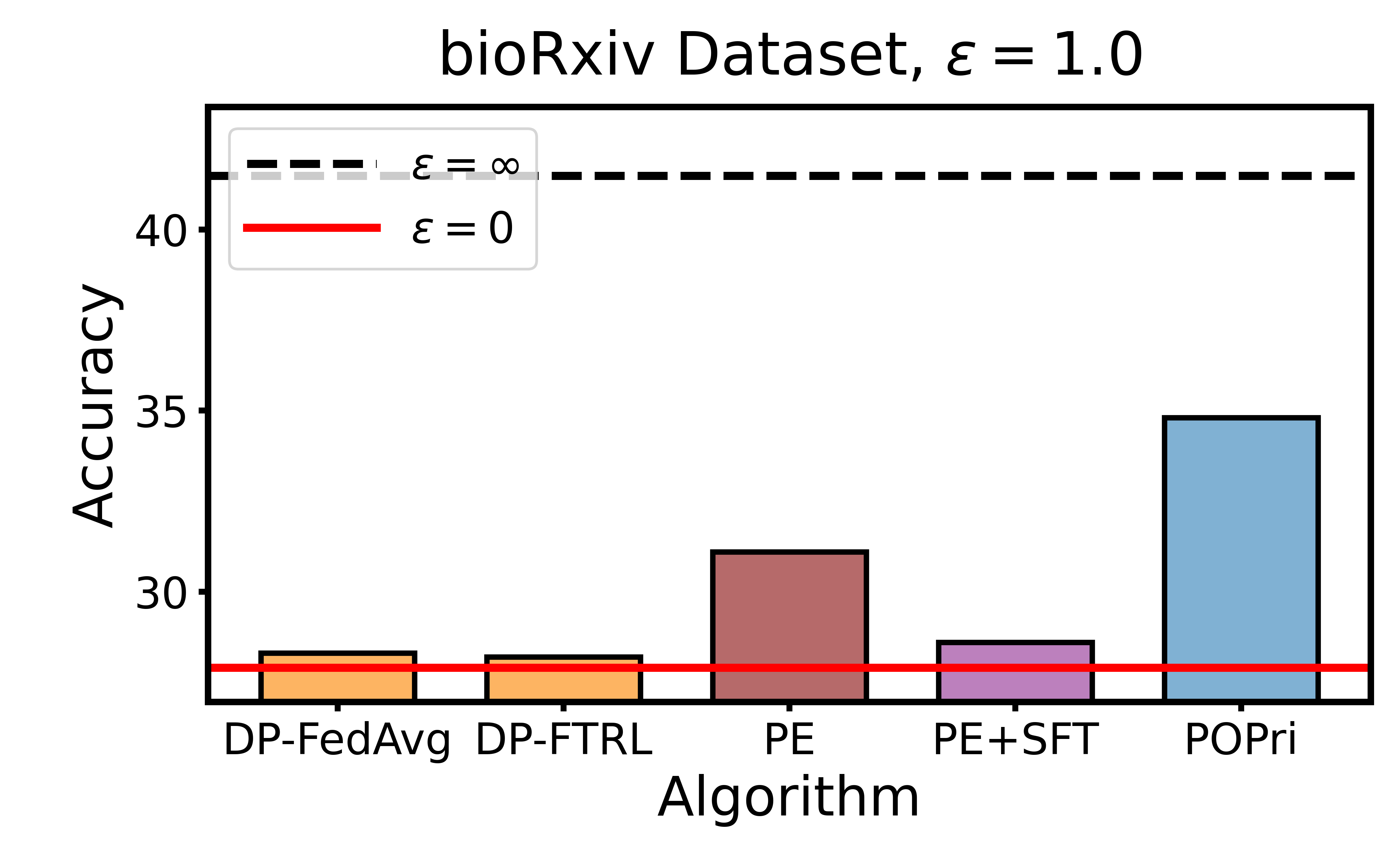}
}
\caption{
\textbf{Left:} Private Evolution (PE)-based techniques. Clients generate low-dimensional statistics which summarize the similarity of the synthetic data to their private samples. These are privately aggregated to refine the synthetic data generation for future iterations. Traditional PE (brown) uses a prompt-based method. \name (blue) improves a naive fine-tuning method (PE+SFT, purple) by fine-tuning the LLM using \emph{policy optimization} rather than fine-tuning directly on aggregated client feedback. \textbf{Right:} Next-token prediction accuracy on the bioRxiv dataset at privacy level  $\epsilon=1$. \name closes the accuracy gap between the fully-private and non-private settings by 58\%, compared to 23\% for prior synthetic data methods, and 3\% for DP federated learning methods.}
\label{fig:alignfl}
\end{figure*}

\section{Introduction}
\label{Intro} 
Many important machine learning (ML) applications feature sensitive datasets that are distributed across client devices (e.g. mobile devices).  
Such ML models are often hosted on client devices.
These \emph{on-device} models offer privacy, latency, and storage benefits relative to centrally-hosted models. Examples include Google's GBoard \citep{hard2019federatedlearningmobilekeyboard, xu2023federatedlearninggboardlanguage, wu2024prompt} and Apple's mobile automatic speech recognition system \citep{paulik2021federatedevaluationtuningondevice}. 
Today, federated learning (FL) is the most widely-used approach in practice for learning on-device models; it trains models locally on user devices and aggregates model updates on a central server \cite{mcmahan2017learning}. FL protects the privacy of client data in part by adopting differentially private (DP) \cite{dwork2006differential} optimization techniques, a combination we refer to as DP-FL \cite{mcmahan2017learning,kairouz2021advances,nguyen2022federated,xu2023learning}. 

With breakthroughs in large language model (LLM) capabilities \citep{anil2023palm, team2023gemini, achiam2023gpt, guo2025deepseek} several research teams have used LLMs to better train models on private client data. 
A common strategy applies standard optimization algorithms (e.g., DP stochastic gradient descent, DP-SGD \citep{abadi2016deep}) to fine-tune models on private client data \citep{kurakin2023harnessing, charles2024fine}. 
These approaches have an important limitation in the on-device setting: most LLMs today are too large to fit on client devices, let alone train on them \cite{radford2019language,touvron2023llama}.

To sidestep the size issue, \citet{wu2024prompt, PrE-Text} view the problem of learning from distributed, private client data (partially) as a DP synthetic data problem. 
These approaches use LLM-assisted workflows to generate privacy-preserving synthetic data, similar to client data, at the server; then they train the on-device model \emph{at the server} on the synthetic data. 
This 
avoids storing the LLM  on client devices. 

In more detail, \citet{wu2024prompt} use  prior public information about the clients to create LLM-generated synthetic data for pretraining. 
For example, for their Google GBoard virtual keyboard application, they use prompts like ``Imagine you are a [GENDER] of age [AGE]. Write some examples of chat messages.'' to generate synthetic samples. 
This prompt was designed entirely using prior qualitative information about the data on client devices.
However, prior information may not always be available.
Moreover, this prompt was not refined based on clients' realized data, which could limit the relevance of the resulting synthetic data.

PrE-Text \citep{PrE-Text} instead uses Private Evolution (PE) \citep{lin2023differentially, xie2024differentially, lin2025differentiallyprivatesyntheticdata} to learn prompts that are relevant to client data. 
PE  iteratively sends synthetic data samples to clients for feedback; each client privately measures the closeness of synthetic samples to their own data, discarding irrelevant samples. 
It returns this feedback to the central server, which crafts a new prompt based on the most relevant synthetic samples.
Finally, an LLM  uses the generated synthetic data to fine-tune a downstream model. 
This method of utilizing LLMs for on-device learning has some shortcomings: 
(1) it relies entirely on \emph{prompting} to teach the LLM to generate relevant synthetic data, which may not be as effective as fine-tuning the weights. 
(2) It discards irrelevant samples, which may themselves contain valuable information, as shown in
reinforcement learning with human feedback (RLHF) \citep{ouyang2022training}.

In this paper, we demonstrate how to better utilize LLMs for on-device learning: we propose \textbf{\name} (Policy Optimization for Private Data), an algorithm that reformulates synthetic data-based approaches for private learning as an LLM policy optimization problem. In~\name, we directly fine-tune an LLM's weights to improve the (DP-noised) similarity scores between generated synthetic data and private client samples. 
The fine-tuned LLM is used to generate synthetic data, which is used to train a downstream model.

\paragraph{Contributions.} In summary, our contributions are:

(1) We propose~\name, a novel method that casts private learning under the synthetic data framework as an LLM policy optimization problem. 
Prior work in this space relied on PE, which uses client feedback exclusively to generate new prompts \cite{PrE-Text,xie2024differentially}. 
We alter this feedback to instead provide client \emph{rewards}, and subsequently exploit recent advances in policy optimization \cite{DPO}. 
This recasting  allows us to more effectively exploit the capabilities of LLMs for on-device learning problems. 

(2) We create and maintain \benchname, a new uncontaminated benchmark of federated client data separated by client for the era of LLMs. The datasets in this benchmark consist of: (1) congressional records in English-speaking countries, and (2) abstracts from bioRxiv, collected starting in April 2023. To our knowledge, this is the first dataset that provides researchers with both (a) over 1,000 clients (congressional records contains 134k clients and bioRxiv contains 57k as of August 2024), and (b) regular updates, allowing researchers to easily filter data to avoid contaminated evaluations \citep{magar2022datacontaminationmemorizationexploitation, zhou2023dontmakellmevaluation, yang2023rethinkingbenchmarkcontaminationlanguage, roberts2023datacontaminationlenstime}. 

(3) We demonstrate the utility of \name on this new benchmark set of datasets, as well as two central (i.e., the setting where all data is present on the server and no server-client communication is needed) DP benchmarks  from prior work \citep{yu2023training, xie2024differentially}.
Across all datasets and tasks (we consider next token prediction and text classification), \name achieves the best downstream metrics. For example, Figure \ref{fig:alignfl} shows that on our bioRxiv dataset at a privacy level of $\epsilon=1.0$, \name outperforms PE-based algorithms by 6 full percentage points, and closes the gap between fully private and non-private baselines by over 58\%, compared to 23\% for PE. It outperforms DP-FL-based methods by even more.
Additional experimental details, results, and ablations are provided in Section \ref{sec:eval}.

\section{Problem Statement and Background}
\subsection{Problem Statement}
We consider a set $\mathcal S$ of  clients, $\mathcal S=\{S_1,\ldots, S_n\}$, where $S_i=\{s_1^{(i)}, \ldots, s^{(i)}_{m_i}\}$ denotes the private text data of client $i\in [n]$, and $m_i$ denotes the number of text samples held by client $i$. We consider the partial participation setting, where only a subset of clients can participate in communication with the server at any point in time \citep{kairouz2021practical, mcmahan2017learning}, which is consistent with practical private on-device learning deployments. We assume $L$ clients participate in each round $t\leq T$ and denote this set $\mathcal S^t$.
We do not assume an \emph{a priori} upper bound on $m_i$.
A central server is given a pre-trained downstream model $\Phi$, which it wants to align with the private client data $\mathcal S$. 
We call the aligned downstream model $\tilde \Phi$.
In the process of learning $\tilde \Phi$, the server may make use of a pre-trained public LLM $\Psi$. 
We observe that $\Psi$ and $\Phi$ are different models in general; 
we will assume the server has access to the weights of both $\Phi$ and $\Psi$. 
The server is subject to two restrictions: (1) client data cannot leave client devices, and (2) the final model $\tilde \Phi$ must protect user-level differential privacy (DP):

{\bf User-level (distributed) differential privacy (DP).} We say two datasets $\mathcal S$ and $\mathcal S'$ are \emph{neighboring} if they differ in at most one client's data. That is, there exists an $i\in [n]$ such that for all $j\neq i$, $S_j = S'_j$. A randomized mechanism $\mathcal{M}$ is ($\epsilon$, $\delta$)-DP if, for any pair of neighboring datasets $\mathcal{S}$, $\mathcal{S^{'}}$ that differ by an entire client's data and any possible output set $E$, it holds that ${\rm Pr}[\mathcal{M}(\mathcal{S}) \in E] \le e^{\epsilon} {\rm Pr} [\mathcal{M}(\mathcal{S^{'}})  \in E] + \delta$. The post-processing property of a DP mechanism ensures that any data-independent transformation applied to its output preserves the same DP guarantees. \cite{dwork2006differential,Dwork2014book}.

We also evaluate on central DP baselines, so we define central DP below; in that case the final model $\tilde{\Phi}$ should protect central DP:

{\bf Central (example-level) differential privacy (DP).} We say two datasets (both fully present on the server) $S = \{s_1, \ldots, s_{m}\}$ and $S' = \{s_1', \ldots, s_{m}'\}$ are \emph{neighboring} if they differ in at most one example's data. That is, there exists an $i\in [m]$ such that for all $j\neq i$, $s_j = s'_j$. A randomized mechanism $\mathcal{M}$ is ($\epsilon$, $\delta$)-DP if, for any pair of neighboring datasets $\mathcal{S}$, $\mathcal{S^{'}}$ that differ by one sample and any possible output set $E$, it holds that ${\rm Pr}[\mathcal{M}(S) \in E] \le e^{\epsilon} {\rm Pr} [\mathcal{M}(S')  \in E] + \delta$.

\paragraph{Goal.}
The server seeks an algorithm to optimize the downstream performance (in our paper, this is either next token prediction accuracy or text classification accuracy) of $\tilde \Phi$ on a test set of private data, subject to an $(\epsilon,\delta)$-DP constraint.

\subsection{Related Work}
There are two main approaches for learning on private data. 
\paragraph{DP optimization-based approaches.}
In natural language processing (NLP) tasks with privacy constraints, DP optimization algorithms (e.g., DP-SGD \cite{abadi2016deep}) are often used to fine-tune massively pretrained LLMs on private data \cite{Bommasani2019,kurakin2023harnessing,charles2024fine}. 
However, in settings where client data cannot leave client devices due to privacy concerns, 
central servers cannot conduct this private fine-tuning. 

An alternative approach is to train models directly \emph{on client devices},  using a server to coordinate information exchange between clients; 
in DP federated learning (DP-FL) \cite{mcmahan2017learning,kairouz2021practical},  (small) model weights are iteratively sent to clients for on-device DP optimization. 
DP-FL has struggled to keep up with the growing  size of LLMs;
many LLMs cannot be stored or trained on client devices \cite{collins2023profit}. 
Recent work explores how to train LLMs in the DP-FL framework. 
Proposed approaches include training only subsets of parameters \cite{charles2023towards}, as well as memory-efficient zero-order optimization \cite{zhang2024dpzero,malladi2023fine}. 
However, these methods still require the storage of the entire model on-device, limiting their practicality.

\paragraph{Synthetic data-based approaches.}
An alternative approach to DP optimization involves generating private synthetic data using LLMs, 
followed by directly fine-tuning downstream models. 
Synthetic data can be generated on the server side, which  bypasses client-side hardware constraints. The post-processing property of DP also implies that DP synthetic data can be used repeatedly without incurring additional privacy loss \cite{yue-etal-2023-synthetic}. In the centralized DP setting (where the server is trusted to gather all the data, as opposed to our private on-device setting), prior studies have shown that training downstream models on DP synthetic text achieves performance comparable to privately training on real data \citep{yue-etal-2023-synthetic,mattern-etal-2022-differentially,xie2024differentially}. In the \textbf{private on-device} setting, \citet{PrE-Text} show that fine-tuning a small model on user-level DP synthetic text data on the server side can actually \emph{outperform} DP-FL, with a significant reduction in communication and computation cost. 
Similarly, \citet{wu2024prompt} show that pretraining an FL model on private synthetic data can improve the final outcome of DP-FL. 

One approach for generating synthetic text data is to fine-tune an LLM (with DP-SGD) on private data \citep{kurakin2023harnessing, yu2024privacypreservinginstructionsaligninglarge} and then using the LLM to generate synthetic data.
However, client hardware
constraints render this approach infeasible on-device. 
Recent works have relied instead on privacy-aware prompt engineering to generate synthetic data \cite{wu2024prompt,xie2018differentially,PrE-Text}. 
An important framework by \citet{lin2023differentially,lin2025differentiallyprivatesyntheticdata} called \textbf{Private Evolution} (PE) is the basis for several competitive DP synthetic text algorithms, including Aug-PE \cite{xie2024differentially} and PrE-Text \cite{PrE-Text}. 
Roughly, these algorithms use the public LLM $\Psi$ to generate synthetic data, score each synthetic data according to its closeness to the client data, and discard synthetic data with low scores. The surviving synthetic data are used as in-context examples for $\Psi$ to generate synthetic data.  
In concurrent work to ours, Zou \emph{et al.} extend the PE framework to generate synthetic data from multiple pretrained language models (LMs), and present ``good" and ``bad" responses to the LMs in the next round for in-context learning \cite{zou2025contrastive}.
Private Evolution may sacrifice data quality in two ways:
First, it uses in-context learning, which is often less effective than fine-tuning \citep{mosbach2023few}. Second, discarding low-score synthetic data may lose useful information \citep{ouyang2022training}.
We address both by turning the DP synthetic generation problem into an LLM policy optimization problem.

\section{\name}
\label{sec: methods}
The core idea of \name (\textbf{P}olicy \textbf{O}ptimization for \textbf{Pri}vate Data) is a natural reformulation of private on-device learning from synthetic data as an LLM policy optimization problem, which enables the use of powerful LLM alignment methods like DPO \citep{DPO}. In this section, we detail the \name design principles and algorithm. \name's design is based on two related questions.

\paragraph{1. What client feedback should we collect for fine-tuning?}  Three natural options arise: 

(1) \textit{DP Data.} Clients could directly transmit DP synthetic data samples for fine-tuning, e.g., using a method like DP-Prompt \citep{dpprompt}. DP-Prompt uses an LLM to summarize text at a temperature specified by the desired DP $\epsilon$ level. 
However, DP text cannot be aggregated into a single statistic, which prevents the use of secure aggregation \citep{secagg}; this increases the noise needed to reach a given DP guarantee. 
As such, prior work has shown that DP-Prompt is not competitive with other private on-device learning methods \citep{PrE-Text}. 
We favor aggregation-compatible representations of client data, such as summary statistics or model parameters. 

(2) \textit{DP Model Parameters.} A second  alternative is to send the parameters of either the LLM $\Psi$ or the downstream model $\Phi$ to the client and train on the private samples with DP-SGD \citep{abadi2016deep}. These parameters are compatible with secure aggregation \citep{secagg}, which makes  more efficient use of DP budget. However, $\Psi$ cannot be sent to clients because of client storage constraints. 
Sending $\Phi$ is the DP-FL approach, which is one of our baselines.

(3) \textit{DP Statistics.} 
Finally, we could collect low-dimensional statistics capturing the quality of synthetic data samples.
In PE, the server generates  $K$ synthetic data samples \cite{xie2024differentially,PrE-Text}, and
each client computes a histogram counting how often each of the private samples is closest to one of the $K$ samples.
This $K$-dimensional histogram can be made DP by adding (comparatively) little noise, and it is amenable to secure aggregation  \cite{xie2024differentially,PrE-Text}.
We view such low-dimensional statistics as the most promising option, as they have lower communication and storage costs, and they make better use of the privacy budget. 
\textit{In a departure from PE, we design the low-dimensional statistics collected by POPri to enable building a preference dataset.}
We ask the server to generate $J$ samples from each of $K$ prompts; each client then scores the $K \times J$ samples according to how well they represent the client's data, and the server aggregates the scores for all the synthetic samples.
Using these scores, the server can construct a ``higher scoring response'' and ``lower scoring response'' pair (a ``preference pair'') for each of the $K$ prompts.
The benefit of this new design ties directly to the next question.  

\vspace{-8pt}
\paragraph{2. How should we use client feedback?} \mbox{}\\
\vspace{-8pt}

Given a vector summarizing the quality of synthetic data samples, how should we use it? A few options arise:

(1) \textit{In-Context Learning.} We could use the highest-scoring synthetic samples as in-context examples to prompt the LLM $\Psi$. 
This is the PE approach \citep{PrE-Text, xie2024differentially}. However, in-context learning typically  performs worse than finetuning-based approaches \citep{mosbach2023few}, and we find experimentally that \name outperforms Private Evolution (PE) (\cref{fig:alignfl}, \cref{tb:result}).

\begin{figure}[ht]
\begin{center}
\centerline{\includegraphics[width=8cm]{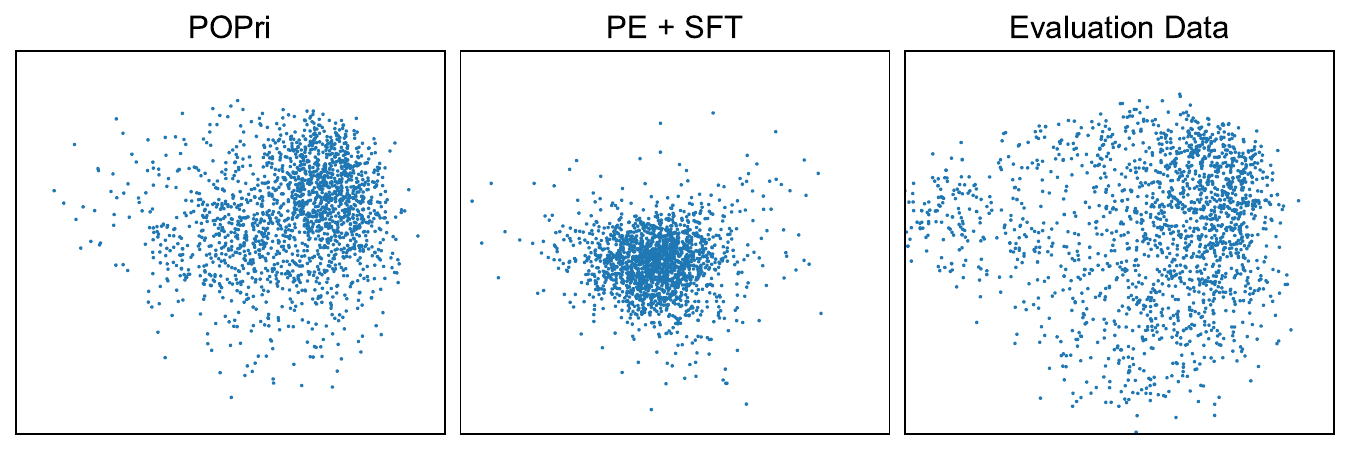}}
\caption{2-PCA visualization of synthetic data from \name and PE+SFT, and evaluation data. We see that POPri's synthetic data distribution (\textbf{left}) is much closer to the evaluation data distribution (\textbf{right}) than the PE+SFT synthetic data distribution (\textbf{middle}). Naive fine-tuning with SFT on PE-generated synthetic data does not make best use of client feedback.}
\label{fig:PCA_comparison}
\end{center}
\vskip -0.25in
\end{figure}

(2) \textit{Supervised Fine-Tuning (SFT).} 
One could directly fine-tune the LLM $\Psi$ on the highest scoring samples using  next-word-prediction loss. 
This is analogous to the SFT baseline evaluated in the RLHF \citep{ouyang2022training} and DPO \citep{DPO} papers, which showed that RLHF and DPO outperform SFT. The reason is that the highest scoring samples--while better than the low-scoring samples--are not perfect responses to the prompt.  
The SFT loss trains the LLM to treat high-scoring samples as perfect responses, which is misaligned with the LLM's task. 
Empirically, we see that this approach (PE+SFT)  produces synthetic data that is not representative of the private data (\cref{fig:PCA_comparison}) and has poor downstream performance (\cref{tb:result}). 

(3) \textit{Policy Optimization (PO).}
Policy optimization-based methods like DPO \cite{DPO} instead directly optimize the LLM to produce higher-scoring samples (where the score can be defined by the user of the algorithm). 
In other words, they are designed to directly make use of the low-dimensional scores we collect from client feedback. 
Hence, we expect such methods to produce higher quality synthetic data, as evaluated on downstream tasks. 

\begin{algorithm}[tb]
   \caption{{\tt \name}}
   \footnotesize
   \label{alg:align-fl}
\begin{algorithmic}[1]
   \STATE {\bfseries Input:} Clients private data $\{S_i\}_{i\in [n]}$, Number of rounds $T$, Number of generated samples $N_{\rm syn}$, Noise multiplier $\sigma$, LLM $\Psi$, embedding model $\Gamma$, base prompt $\eta$, participating clients in each round $\mathcal S^t$, ``rejected" index $\ell$, random prompt generator $\Lambda(\cdot)$, number of clients sampled $L$\\
   \STATE {\bfseries Output:} Synthetic data $S_{syn,T+1}$
   \STATE
   \STATE All clients $i\in [n]$ embed private samples, $E_{i}=\Gamma(S_{i})$ 
   \STATE Server initializes LLM $\Psi_1 = \Psi$
   \FOR{$t\leftarrow1$ $...$ $T$ }
   \STATE {\bfseries \underline{Server}:}
   \STATE Initialize the response vector $R=\emptyset$
   \FOR{$k\leftarrow 1 \ldots K$ }
   \STATE Generate prompt $\eta_k=\Lambda(\eta)$, 
   \STATE \colorbox{SkyBlue}{Generate $J$ responses  $R_{kj} = \Psi_t(\eta_k)$, $j\in [J]$}
   \ENDFOR
   \STATE Send embeddings $E_{syn,t}=\{\Gamma(R_{kj})\}_{k\in [K], j
   \in [J]}$ to all clients in $\mathcal S^t$
   \STATE
   \STATE {\bfseries \underline{Client $i\in \mathcal S^t$}:}
   
   \STATE \colorbox{SkyBlue}{${\rm Scores}_{i,t} \leftarrow$ ${\tt SIMILARITY}(E_{syn,t}, E_{i})$}
   \STATE Send ${\rm Scores}_{i,t} + \mathcal{N}(0, \sigma^2 I / L)$ to Server
   \STATE
   \STATE {\bfseries \underline{Server}:}
   \STATE Secure aggregate scores: ${\rm Scores}_{t} = \frac{1}{L} \sum_{i\in \mathcal{S}^t} {\rm Scores}_{i,t}$
   \STATE Set $P[k,j]$ as the $j$-th highest score response for prompt $\eta_k$, according to  ${\rm Scores}_{t}$
   \STATE Initialize preference dataset $\mathcal{P}_t = \emptyset$
   \FOR{$k\leftarrow 1 \ldots K$ }
   \STATE \colorbox{SkyBlue}{Select positive synthetic sample: $\mathcal P_{t}[k,1] = P_t[k,1]$}
   \STATE \colorbox{SkyBlue}{Select negative synthetic sample: $\mathcal P_{t}[k,2] = P_t[k,\ell]$}
   \ENDFOR
   \STATE \colorbox{SkyBlue}{Fine-tune: $\Psi_{t+1} \leftarrow \text{DPO}(\Psi_{t}, \{\eta_k\}_{k\in[K]}, \mathcal P_{t})$ }
   \ENDFOR
   \STATE {\bfseries \underline{Server}:}
   \STATE
   Output final synthetic data $S_{syn,T+1}$ from $\Psi_T$

\end{algorithmic}
\label{alg:alignfl}
\end{algorithm}

\subsection{\name Algorithm}
Pseudocode can be found in \cref{alg:alignfl}. We highlight the algorithmically new steps (that differ from PE) in \colorbox{SkyBlue}{blue}.

\paragraph{\colorbox{SkyBlue}{1. Synthetic sample generation.}} We generate $K$ prompts (details in \cref{app:prompt}). A prompt is generated by randomly sampling three samples from $\Omega$ and prompting LLaMA-3-8B \citep{touvron2023llama} to generate a fourth sample given the first three samples as examples. The exact prompt is given in \cref{sec:align-FL implementation}. For each of the $K$ prompts, we generate $J$ synthetic samples (by running the prompt independently $J$ times). In total, the server generates $K \times J$ synthetic samples, embeds them using a small sentence embedding model $\Gamma$ and sends the embeddings to every client in $\mathcal S^t$, i.e., the clients sampled in round $t$.

\paragraph{\colorbox{SkyBlue}{2. Scoring synthetic data using DP client feedback.}} 
Next, each client in $\mathcal{S}^t$ scores the synthetic samples. Specifically, each client calculates, for each of the $K \times J$ synthetic samples, its  cosine similarity with each of the client's private samples, averaged over the client's samples (\cref{alg:dpnn}).
The use of cosine similarity differs from PE, which uses a nearest neighbors histogram \citep{lin2023differentially, PrE-Text, xie2024differentially}--using cosine similarity is critical to the performance of POPri as we found in our ablations (see \cref{sec:ablations}).
These similarities for every synthetic sample are arranged into a vector. We clip this vector to a norm of 1, which caps the contribution of each client (similar to how gradient updates are clipped per client in DP-FL \citep{mcmahan2017learning}). Clipping is done primarily for privacy reasons, as we will elaborate later. Clipping also ensures that the contribution of clients with large amounts of data does not overwhelm the contribution of clients with small amounts of data.  
We then add $\mathcal{N}(0, \sigma^2 I / L)$ (where $I$ is the identity matrix of size $KJ \times KJ$) noise to the resulting vector 
to ensure DP ($\sigma^2$ controls the $(\epsilon, \delta)$).
Finally, we aggregate scores via secure aggregation \citep{secagg}, yielding a DP score for each synthetic sample that reflects its relevance to client data.

\begin{table*}[ht]
\footnotesize
\caption{Accuracy ($\%, \uparrow$) of different algorithms on a variety of tasks and datasets (bioRxiv, Congress, PubMed are next-token-prediction accuracy, OpenReview is text classification accuracy). The highest accuracy across all methods is in ${\bf bold}$. All standard deviation error bars are less than 0.5.}

\label{tb:result}
\scriptsize
\vskip 0.15in
\begin{center}
\begin{tabular}{lccccccc}
\toprule
\textbf{Dataset} & \textbf{Method} & \textbf{Data Type} & \textbf{On-device Model}& \textbf{\bf $\epsilon = \infty$}&\textbf{\bf $\epsilon = 7$} & \textbf{$\epsilon = 1$} & \textbf{\bf $\epsilon = 0$} \\
\midrule
 \multirow{5}{*}{bioRxiv } &  DP-FedAvg &Original  & \multirow{5}{*}{DistilGPT2}& \multirow{5}{*}{41.5}& 29.0 & 28.3&\multirow{5}{*}{27.9} \\
& DP-FTRL & Original&  &  &29.0 & 28.2\\
& PE & Synthetic& & & 31.0& 31.1\\
& PE + SFT & Synthetic& & & 28.6 & 28.6\\
& \name (ours) & Synthetic &  & &\textbf{34.4} & \textbf{34.8}\\
\midrule
\multirow{5}{*}{Congress}  &  DP-FedAvg & Original & \multirow{5}{*}{DistilGPT2}& \multirow{5}{*}{35.7} &29.1& 29.0 & \multirow{5}{*}{26.9}\\
& DP-FTRL & Original& & & 29.1&29.0&\\
& PE & Synthetic& & & 27.3 & 27.0 \\
& PE + SFT & Synthetic& & & 27.1 & 27.1 \\
& \name (ours) &Synthetic & & &\textbf{30.6}& \textbf{30.4}\\
\midrule
\multirow{2}{*}{PubMed \citep{yue2023synthetictextgenerationdifferential}} 
& PE & Llama-2-7b-chat-hf, Synthetic (2000)&\multirow{2}{*}{${\textsc{BERT}_{\rm small}}$}  &\multirow{2}{*}{47.6}&--- & 27.5\\
& PE & Opt-6.7b, Synthetic (2000)& & &--- & 27.9\\
& \name (ours) & Synthetic (2000)& & & --- & \textbf{29.4}\\
\midrule
\multirow{2}{*}{OpenReview \citep{xie2024differentially}} 
& PE & Llama-2-7b-chat-hf, Synthetic (2000)&\multirow{2}{*}{${\textsc{RoBERTa}_{\rm base}}$}  &\multirow{2}{*}{50.8}&--- & 37.0 & \multirow{2}{*}{32.0} \\
& PE & Opt-6.7b, Synthetic (2000)& & &--- & 32.1 & \\
& \name (ours) & Synthetic (2000)& & & --- & \textbf{40.2} & \\
\bottomrule
\end{tabular}
\end{center}
\vskip -0.18in
\end{table*}
\vspace{-10pt}

\paragraph{\colorbox{SkyBlue}{3. LLM Policy Optimization.}} The key insight of our paper is that by generating $J$ synthetic samples from $K$ prompts and scoring all of them using DP client feedback, we can create a preference dataset where for each of the $K$ prompts, we can assemble a ``good sample'' and a ``bad sample''. This design choice allows the usage of powerful LLM policy optimization algorithms (we choose DPO \citep{DPO}) to finetune the LLM $\Psi$.
In detail, each of the $K$ prompts have $J$ synthetic samples which are ranked according to the scores we gathered. Then for each of the $K$ prompts, we set the highest scoring sample as the ``chosen sample'' and the $\ell$-th highest scoring sample as the ``rejected sample''. This resulting preference dataset can then be passed, along with the LLM $\Psi$, into the DPO preference optimization loss \citep{DPO}:
\begin{align*}
    \label{eq:dpo}
    \min_{\Psi} \underset{\underset{y_r}{x, y_\omega}}{\mathbb E}\left[-\log s\left(\tau\log(\frac{\Psi(y_\omega|x)}{\Psi(y_r|x)})- \tau\log(\frac{\Psi_{\text{ref}}(y_\omega|x)}{\Psi_{\text{ref}}(y_r|x)})\right)\right]
\end{align*}
where $\Psi_\text{ref}$ a fixed checkpoint for the LLM (we use the public checkpoint of the LLM), $\tau$ is a parameter controlling deviation of $\Psi$ from $\Psi_\text{ref}$, $x$ is the prompt, $y_\omega$ is the chosen sample, $y_r$ is the rejected sample, $\Psi(y|x)$ is the probability of generating $y$ given $x$ for $\Psi$, and $s$ is the sigmoid function. The expectation is taken with respect to the empirical distribution (i.e. real samples).
The DPO loss will be used to finetune $\Psi$ to generate more samples similar to the chosen sample and fewer like the rejected sample. To reduce GPU memory use, we use LoRA \citep{hu2021lora} on all the attention matrices and up/down projection matrices with a rank of 4, $\alpha=8$.  After fine-tuning over the $K$ prompts and preference pairs, we return back to step (2) and generate new synthetic data using the newly fine-tuned $\Psi$.

\paragraph{4. Synthetic data generation for downstream tasks.} Using the final version of $\Psi$, we generate a large set of synthetic data $S_{syn,T+1}$ which is used to fine-tune $\Phi$ into $\tilde{\Phi}$.  $\tilde{\Phi}$ is then sent to all the client devices, where they can perform inference without communicating information to the server.

\paragraph{Privacy guarantees.} Because each client's vector is clipped to 1, and the only information revealed to the server (or any other party) is the aggregated vector, the sensitivity of the algorithm is 1. We add $\mathcal{N}(0, \sigma^2 I/L)$ noise to each client's vector, so the vector given to the server has noise $\mathcal{N}(0, \sigma^2 I)$, satisfying the  Gaussian Mechanism with sensitivity 1. To calculate privacy, we can use a privacy accountant like \texttt{OPACUS.ACCOUNTANTS.ANALYSIS.RDP} \citep{yousefpour2021opacus}, and input $T$ (the number of rounds we run the algorithm, $q$ (the fraction of clients sampled per round), $\delta$, and set $\sigma$ to get the desired $\epsilon$ value.

\section{\benchname: A Federated Benchmark for LLM Evaluation} 
\label{sec:benchmark}

Today, the most widely-used evaluation datasets for federated learning of text models come from the work of \citet{reddi2020adaptive}; they include text from StackOverflow posts and Shakespeare plays.
These datasets pose two evaluation challenges: (1) 
They pre-tokenize  inputs in a non-invertible way, 
which prevents researchers from using custom tokenizers adopted by several LLMs.
(2) The datasets may lead to contaminated evaluations. 
As state-of-the-art LLMs have been trained on large swaths of the public internet, old public benchmark datasets may be in the training data of many LLMs \citep{magar2022datacontaminationmemorizationexploitation, zhou2023dontmakellmevaluation, yang2023rethinkingbenchmarkcontaminationlanguage, roberts2023datacontaminationlenstime}. 
To our knowledge,  one work proposes a benchmark dataset for federated LLMs \citep{ye2024fedllm}. The datasets in this paper have at most 747 clients, which may be insufficient for simulating production use cases.
Further, they do not explicitly avoid contamination.

We release \textbf{\benchname}, a benchmark comprising two new datasets, Congressional Speeches and bioRxiv, for experiments over federated client data. These datasets  (a) allow researchers to easily avoid contamination, and (b) provide enough distinct clients to simulate production settings. 

\begin{figure*}[ht]
\begin{center}
\centerline{\includegraphics[width=8.0cm]{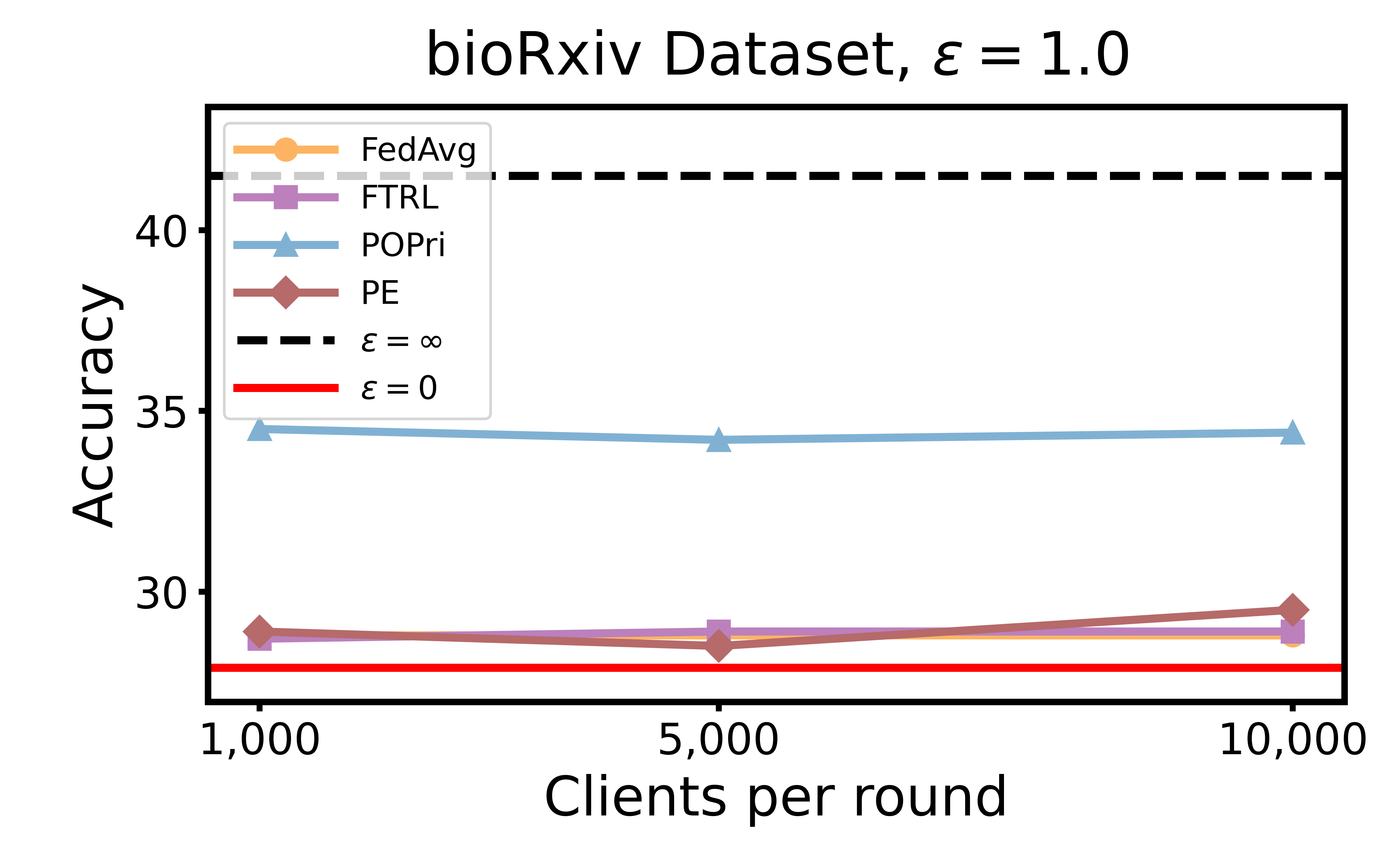}}
\scriptsize
\vskip -0.15in
\caption{Next-token prediction accuracy performance of four methods as a function of the number of clients sampled per round out of 10000. We see that across different client participation scenarios, \name consistently performs the best.
}
\label{fig:scaling-eps1}
\end{center}
\vskip -0.2in
\end{figure*}

\textbf{Congressional Speeches (``Congress")}\footnote{\url{https://huggingface.co/datasets/hazylavender/CongressionalDataset}}  is a dataset of 134k speeches or debates 
scraped from congressional or parliamentary transcripts in the US, UK, and Canada. 
We treat each speech as a separate client, and samples are created as successive 64-token spans within the speech. 
\textbf{bioRxiv}\footnote{\url{https://huggingface.co/datasets/hazylavender/biorxiv-abstract}} is a dataset of 57k abstracts, each of which we consider a client dataset of strings, scraped from biology papers. Samples are 64-token spans of the abstract.
More details on the datasets are included in \cref{app:dataset}.

A key feature of our datasets is that they are  updated every 6 months and sorted by date.
Hence, researchers can easily select datasets that were generated after their model's knowledge cutoff date. 
In this paper, we use data from \benchname published between the dates of April 2023 to August 2024 to avoid contamination with the latest LLM we evaluate our algorithms with, LLaMA-3-8B \citep{llama-model-card}--which has a knowledge cutoff of March 2023.

\section{Experiments}
\label{sec:eval}

\paragraph{Datasets and tasks.}
For next token prediction accuracy, we evaluate \name on the \benchname datasets (\textbf{Congress} and \textbf{bioRxiv}),
as well as \textbf{PubMed} \citep{yu2023training, xie2024differentially} used in the evaluation of Private Evolution (Aug-PE) \citep{xie2024differentially}.  
PubMed contains abstracts of medical papers published between August 1-7, 2023 (details in \cref{app:pubmed}). For text classification, we evaluate \name on \textbf{OpenReview} consisting of ICLR 2023 reviews published on November 5, 2022 which was used in the evaluation of Private Evolution (Aug-PE) \citep{xie2024differentially}. Note that PubMed and OpenReview are evaluations in the \textit{central DP setting}, where the entire dataset is present on the server and no server-client communication is needed. To execute \name on PubMed, we use the central DP version of \name, detailed in \cref{alg:popri-central}. For OpenReview, we employ conditional generation\footnote{Simply speaking, we score synthetic data generated for each class separately but combine the preference datasets into one to finetune the LLM $\Psi$.} (similar to PE \citep{lin2023differentially, xie2024differentially}, where the generation is conditioned on being given a class to generate. The (central) conditional generation version of \name is detailed in \cref{alg:popri-conditional}.

\paragraph{Models.} 

\textit{Next token prediction tasks:} We use LLaMA-3-8B for the LLM $\Psi$
\cite{grattafiori2024llama3herdmodels}, which has a knowledge cutoff date of March 2023 \cite{llama-model-card}.
For embedding models (used in measuring semantic distance between text samples), we use `all-MiniLM-L6-v2' sentence transformer \cite{reimers-gurevych-2019-sentence}. We choose DistilGPT2 \citep{sanh2019distilbert} as the downstream on-device language model for the \benchname evaluations, which has only 82M parameters, and $\text{BERT}_{\text{small}}$ as the downstream model for the PubMed evaluation to be consistent with \citet{xie2024differentially}.
For synthetic text generation (using the LLM $\Psi$), we set the maximum sequence length to 64 for the bioRxiv and Congressional Speeches evaluations and 512 for PubMed/OpenReview. 

\textit{Text classification:} We use LLaMA-2-7b-chat-hf for the LLM $\Psi$ \citep{touvron2023llama} to ensure our evaluation was not contaminated, as the knowledge cutoff for LLaMA-2-7b-chat-hf is September 2022 (before the publish date of the ICLR 2023 reviews). For the embedding model we use the `sentence-t5-xl' sentence transformer \cite{reimers-2019-sentence-bert}, and use $\text{RoBERTa}_{\text{base}}$ as the downstream model to be consistent with \citet{xie2024differentially}.

\paragraph{Metrics.} We primarily evaluate each method on accuracy (next-token or text classification) of the final downstream on-device model $\tilde{\Phi}$. In some ablations we also measure the distance of the synthetic dataset to the private dataset using the Fréchet Inception Distance (FID) \citep{heusel2017gans}. During training, we evaluate the models on the validation dataset and select the checkpoint that achieves the best validation performance as the model that is evaluated on the test set.

\paragraph{Baselines.} 
We compare \name to several baselines:
(1) DP-FedAvg \cite{pmlr-v54-mcmahan17a} (2) DP-FTRL \cite{kairouz2021practical} (3) Private Evolution (PrE-Text \citep{PrE-Text} and Aug-PE \citep{xie2024differentially}). DP-FedAvg and DP-FTRL directly privately fine-tune the downstream model $\Phi$ on the client data. Private Evolution (PrE-Text and Aug-PE) generates synthetic data on which the downstream on-device model $\Phi$ is finetuned. 
Note that on the PubMed and OpenReview dataset, we compare to Aug-PE results  obtained with models of similar size to the model we use (7B-8B parameters) and which are not potentially contaminated (i.e. model was possibly trained on the benchmark dataset).
We also include $\epsilon = 0$ (fully private) and $\epsilon = \infty$ (fully non-private) baselines. The $\epsilon=0$ baseline for the \benchname evaluations evaluates the public DistilGPT2 checkpoint on the test sets with no further fine-tuning. The $\epsilon = \infty$ baseline is the downstream model finetuned directly on the private training set centralized on the server with no noise. The $\epsilon=0$ baseline for OpenReview is the accuracy obtained by predicting everything to be the most populous class. More details about the setup can be found in \cref{sec:baseline implementation,app:dataset-eval}.

\begin{figure*}[ht]
\begin{center}
\centerline{\includegraphics[width=12.7cm]{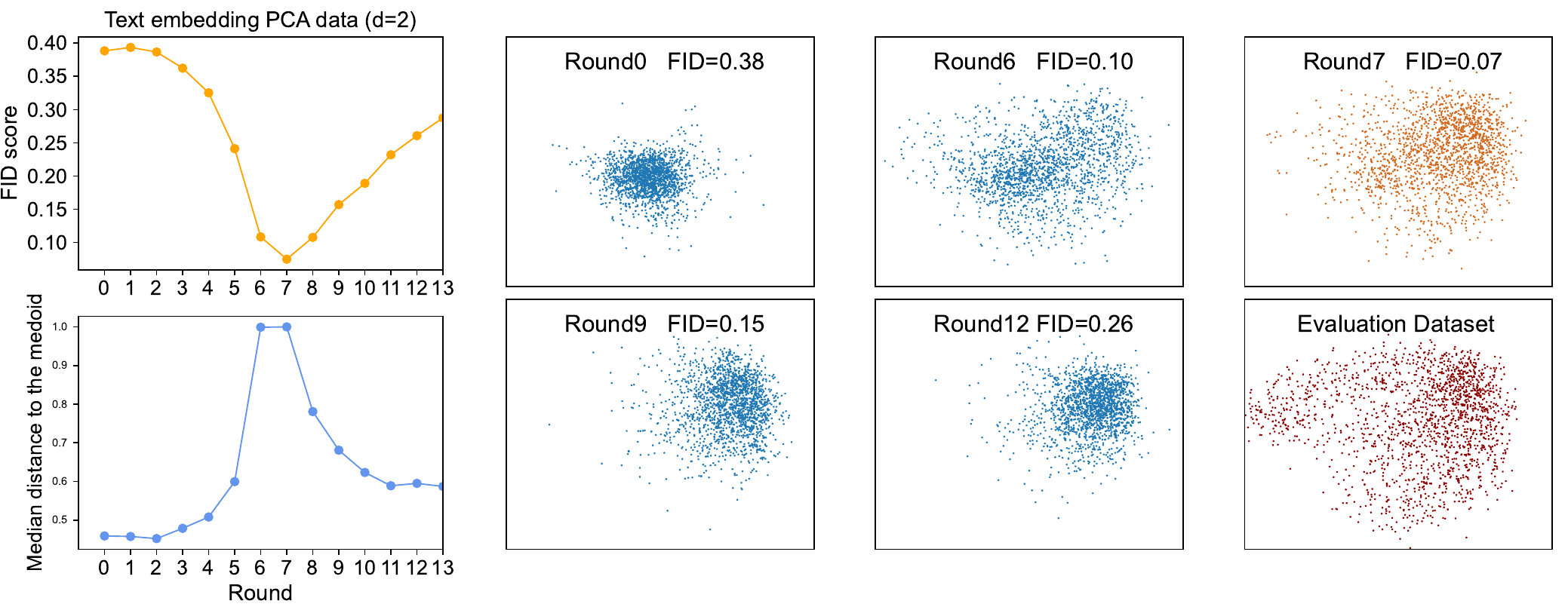}}
\scriptsize
\vskip -0.15in
\caption{PCA visualization of \name synthetic data embeddings over rounds. \textbf{Right 6 Panels:} PCA-2 plots for synthetic data and evaluation data embeddings from the best checkpoint each round for 20 iterations. The orange (round 7) and maroon point clouds represent the round with the lowest FID score and the validation dataset, respectively. \textbf{Top Left Panel:} FID score vs. rounds. \textbf{Bottom Left Panel:} Median distance to the medoid vs rounds. 
Running \name for too many rounds appears to cause overfitting.
}
\vskip -0.2in
\label{fig:PCA}
\end{center}
\end{figure*}

\begin{table*}[ht]
\footnotesize
\begin{center}
\begin{tabular}{lcccc}
\toprule
\textbf{Method} & \textbf{Download} & \textbf{Upload} & \textbf{Client Runtime} & \textbf{Server Runtime} \\
 & \textbf{(floats)} & \textbf{(floats)} & \textbf{(GPU sec)} & \textbf{(GPU sec)} \\
\midrule
FedAvg & 82 million & 82 million & 4.8 & -- \\
PE & 700,000 & 1,800 & 0.0027 & 326.25 \\
\name & 7 million & 18,000 & 0.01 & 13,547.84 \\
\midrule
Reduction factor (FedAvg / \name) &
\textcolor{bettergreen}{$11.71 \times$} &
\textcolor{bettergreen}{$4555 \times$} &
\textcolor{bettergreen}{$480.0 \times$} &
-- \\
Reduction factor (PE / \name) &
\textcolor{red}{$0.100 \times$} &
\textcolor{red}{$0.100 \times$} &
\textcolor{red}{$0.270 \times$} &
\textcolor{red}{$0.024 \times$} \\
\bottomrule
\end{tabular}
\end{center}
\caption{\textbf{Table setting.} Communication and computation cost comparison per round (and per-client for download/upload/client runtime cost) across methods on the bioRxiv dataset with 1000 clients sampled per round. Download and upload are measured in floats; runtimes are measured in GPU seconds (lower is better). ``Reduction factor ($X$ / \name)'' is the cost of method $X$ divided by the cost of \name for the given resource; \textcolor{bettergreen}{green} is a reduction, \textcolor{red}{red} is an increase. Server runtime for FedAvg is left blank as it is negligible compared to other methods.
Overall, we view \name as suitable when server compute is relatively cheap, and improved sample quality is important enough to justify higher on-device communication and computation costs relative to PE (\cref{tb:result}).
}
\label{table:comm_runtime_comparison}
\end{table*}
\paragraph{Privacy Analysis.} 

All baselines use a privacy guarantee of ($\epsilon$, $\delta$)-DP where $\delta$=3$\times 10^{-6}$ and $\epsilon$=1 or $\epsilon$=7 for each of the bioRxiv and Congressional Speeches datasets. 
For PubMed/OpenReview, we set $\delta < {1 \over N_{priv}\cdot {\rm log}(N_{priv})}$ ($N_{priv}$ is the number of private samples).
We follow the privacy accounting method detailed in \cref{sec: methods} for \name. 
Details for all baselines are in \cref{app:privacy-details}.

\subsection{Main Results}
Table~\ref{tb:result} lists the accuracy (next token prediction and text classification) achieved by baseline methods (DP-FedAvg, DP-FTRL, Private Evolution) and \name. 
In this table, we assume full participation (no client sampling) for fair comparison to baselines, some of which do not have client sampling versions.
We find that \name outperforms all the baseline algorithms. Furthermore, in the $\epsilon = 1$ setting \name closes the gap between fully private learning ($\epsilon=0$) and fully non-private learning ($\epsilon=\infty$) by 40-58\% depending on the setting, compared to PE which closes 1-28\%.
For all methods tested, the measured accuracy values do not depend strongly on $\epsilon$. This has been observed in prior work on DP synthetic data using LLMs \cite{xie2024differentially,PrE-Text}. \name outperforms Private Evolution (Aug-PE) even when holding our synthetic sample budget to 2000. Note that synthetic samples are cheap in \name (we could generate many more) because we have access to the full model, while \citet{xie2024differentially} only assume access to a model API.

\paragraph{Cost analysis case study.} 
In \cref{table:comm_runtime_comparison} we analyze the per-round communication and computation costs (and per-client, for download/upload/client runtime costs) of FedAvg (a representative and cheap method among the DP-FL-based methods), PE, and POPri on the bioRxiv dataset experiment with 1000 clients sampled per round.

For FedAvg, each round the sampled clients download and upload the downstream model, which in our case is DistilGPT2. This is an 82M (82 million) parameter model leading to a download and upload cost of 82M floats. The client runtime cost comes from local gradient computation, and server runtime is negligible because the server only needs to average model deltas from the clients. For PE, the communication cost comes from each client downloading $K = 1800$ sentence embeddings of size 384 resulting in a download cost of 700K (700,000) floats, and uploading a histogram of size 1800 resulting in an upload cost of 1800 floats.  The client runtime cost comes from calculating a nearest neighbors histogram and the server runtime cost for PE comes mainly from using the LLM $\Psi$ to generate synthetic samples each round. In POPri each client downloads $K \times J = 1800 \times 10$ sentence embeddings for a download cost of 7M floats and uploads a vector of size 18,000 for an upload cost of 18,000 floats. The client runtime cost of POPri comes from calculating the cosine similarities, and the server runtime comes from both using $\Psi$ to generate synthetic samples and running DPO.

\textbf{Interpretation.} In summary, \name is much more communication-efficient and client compute-efficient than FedAvg, while using much more server compute. On the other hand, \name is generally more communication- and computationally-expensive than PE. At the same time, \name has the best downstream performance among all three methods, as seen in \cref{tb:result}. 
Hence, \name can be a suitable method when (1) server compute is cheap and powerful, and (2) getting the best synthetic data/downstream model quality is important.

\subsection{Ablations}
\label{sec:ablations}
\paragraph{Cosine similarity vs. Nearest neighbors histogram.} Private Evolution \citep{lin2023differentially, PrE-Text, xie2024differentially} uses a DP nearest neighbors histogram calculation to score the quality of synthetic samples. The DP nearest neighbors histogram sets the score of a particular synthetic sample to the number of private samples that are closest to that particular synthetic sample (under some text embedding function). In \name, we instead set the score of a particular synthetic sample to the average cosine similarity between that particular synthetic sample and all private samples (under some text embedding function). We find that cosine similarity works  much better than a nearest neighbors histogram (\cref{fig:cosine_vs_nnhist}), possibly because nearest neighbor histograms produce sparser scores, often assigning zero to all synthetic samples associated with a given prompt.
In this setting, the chosen and rejected samples for preference optimization end up being essentially random. In contrast, cosine similarity provides denser scoring that allows the construction of meaningful preference pairs for all prompts.

\paragraph{Partial client participation.}
In each round a fixed number of clients is subsampled uniformly at random for feedback generation.
\cref{fig:scaling-eps1} 
shows the next-token prediction accuracy ($\%$) of four algorithms for different numbers of clients per round.   
\name  consistently outperforms all of the baselines, regardless of the client sampling rate. Moreover, \name's accuracy is not sensitive to the client sampling rate. 

\paragraph{Data Distribution Evolution.}
Synthetic datasets are often generated using a language model distinct from the one being aligned \citep{online_DPO}, making the alignment phase inherently off-policy as the model evolves during training. This is reflected in the synthetic data, where the FID score (relative to a held-out evaluation set) worsens after improving. Figure~\ref{fig:PCA} shows PCA visualizations of synthetic data embeddings across alignment iterations, while the left panels plot the FID score and median distance to the medoid in the PCA space. 
The data distribution transitions from being initially clustered to (roughly) matching the true data distribution, back to being clustered, 
likely due to overfitting. Early stopping based on validation metrics can help.


\paragraph{How to select rejected samples.}
Unlike vanilla DPO, we can select the ``chosen'' and ``rejected'' sample pair from the $J$ samples for each of the $K$ prompts. 
We consistently choose the highest-scoring sample (rank 1) as the ``chosen'' sample, but there are different options for the ``rejected'' sample. 
We found that the middle-ranked sample (e.g.,  $\ell=5$th-ranked out of $J=10$) yields the best results, rather than using the last-ranked sample. If the rejected sample is too dissimilar to client data, then the preference pair is uninformative. However, choosing a sample that is too similar to client data (e.g., rank 2) for the rejected sample could lead to incorrect preference pairs due to DP noise swapping rankings. 
We use the 5th-ranked sample, and justify it experimentally in \cref{sec:rejected_sample}.

\section{Conclusion}
Private on-device learning is important when data is stored on edge devices with hardware, storage, and privacy constraints. 
We propose \name, which recasts synthetic data-based approaches for private learning as an LLM policy optimization problem. 
\name makes several novel design choices in how it gathers and utilizes client feedback to generate DP synthetic data, which is used to finetune a downstream on-device model. \name  outperforms DP-FL and synthetic data baselines on a variety of tasks, including on a large-scale \benchname, a new federated benchmark we have curated.

\section*{Impact Statement}
In this paper, we train models satisfying differential privacy guarantees. When using differential privacy as a tool for protecting user data, it is important to communicate to users what the privacy guarantees mean to be able to obtain informed consent. The algorithms in this paper also use LLMs, which were trained on large scale public text data. While this data was public, explicit consent may not have been given for its use in training the models. The algorithms using LLMs in the paper make no claims about the privacy guarantees of data used in the pretraining of the LLMs. 

While our work aims to show how synthetic data can be useful for federated learning, it also poses a number of ethical risks, including the generation of biased or harmful content. In particular, our method (and all variants of private evolution) inherits the biases and undesirable aspects of the public LLM. For example, suppose the public LLM only generates text in English, but some clients’ private data is all in Spanish. In these settings, clients would be forced to vote on synthetic samples, even if potentially none of them are relevant to the client. This may cause the client to contribute data reinforcing a model that is actively not useful (or even harmful) to the client. In contrast, DP-SGD methods do not suffer from this shortcoming, because they do not rely on a public LLM. This problem raises an important point—how can we design DP synthetic data algorithms in which clients can stem the biases or failures of the public LLM, based on their own data? This important question is beyond the scope of the current paper.

\section*{Acknowledgments}
This work was supported in part by NSF grants 	CCF-2338772 and CNS-2148359, as well as C3.ai, Bosch, Intel, and the
Sloan Foundation.
This work used Bridges-2 GPU \citep{br2,Neo_BR2} at the Pittsburgh Supercomputing Center through allocation CIS240135 and CIS240937 from the Advanced Cyberinfrastructure Coordination Ecosystem: Services \& Support (ACCESS) program, which is supported by National Science Foundation grants 2138259, 2138286,
2138307, 2137603, and 2138296 \citep{boerner2023access}. The
authors acknowledge the National Artificial Intelligence Research Resource (NAIRR) Pilot, the AI and Big Data group at the Pittsburgh Supercomputing Center, and NCSA Delta GPU for contributing to this research result. 

\nocite{langley00}

\bibliography{ref}
\bibliographystyle{icml2025}

\newpage
\appendix
\label{appendix}
\onecolumn
\section{Algorithmic Details}
Here we provide the pseudocode for the centralized version of \name and the central + conditional generation version of \name.

\begin{algorithm}[H]
   \caption{{\tt \name} (central DP, unconditional)}
   \footnotesize
   \label{alg:popri-central}
\begin{algorithmic}[1]
   \STATE {\bfseries Input:} Number of iterations $T$, Noise multiplier $\sigma$, LLM $\Psi$, embedding model $\Gamma$, base prompt $\eta$, random prompt generator $\Lambda(\cdot)$, ``rejected" index $\ell$, private dataset $S$, $K$ number of prompts, $J$ number of responses per prompt\\
   \STATE {\bfseries Output:} LLM for generating synthetic data $\Psi_{T + 1}$
   \STATE
   \STATE Embed all private samples $E = \Gamma(S)$ 
   \STATE Initialize LLM $\Psi_1 = \Psi$
   \FOR{$t\leftarrow1$ $...$ $T$ }
   \STATE Initialize the response vector $R=\emptyset$
   \FOR{$k\leftarrow 1 \ldots K$ }
   \STATE Generate prompt $\eta_k=\Lambda(\eta)$, 
   \STATE Generate $J$ responses  $R_{kj} = \Psi_t(\eta_k)$, $j\in [J]$
   \ENDFOR
   \STATE Calculate embeddings $E_{syn,t}=\{\Gamma(R_{kj})\}_{k\in [K], j
   \in [J]}$
   
   \STATE ${\rm Scores}_{t} \leftarrow$ ${\tt CENTRALSCORE}(E_{syn,t}, E) + \mathcal{N}(0, \sigma^2 {I})$
   \STATE Set $P[k,j]$ as the $j$-th highest score response for prompt $\eta_k$, according to  ${\rm Scores}_{t}$
   \STATE Initialize preference dataset $\mathcal{P}_t = \emptyset$
   \FOR{$k\leftarrow 1 \ldots K$ }
   \STATE Select positive synthetic sample: $\mathcal P_{t}[k,1] = P_t[k,1]$
   \STATE Select negative synthetic sample: $\mathcal P_{t}[k,2] = P_t[k,\ell]$
   \ENDFOR
   \STATE Fine-tune: $\Psi_{t+1} \leftarrow \text{DPO}(\Psi_{t}, \{\eta_k\}_{k\in[K]}, \mathcal P_{t})$ 
   \ENDFOR
   \STATE
   Output $\Psi_{T + 1}$

\end{algorithmic}
\end{algorithm}

\begin{algorithm}[H]
   \caption{{\tt \name} (central DP, conditional)}
   \footnotesize
   \label{alg:popri-conditional}
\begin{algorithmic}[1]
   \STATE {\bfseries Input:} Number of iterations $T$, Noise multiplier $\sigma$, LLM $\Psi$, embedding model $\Gamma$, base prompt $\eta$, conditional (class-specified) random prompt generator $\Lambda(\cdot, \cdot)$, ``rejected" index $\ell$, private dataset $S$, $K$ number of prompts, $J$ number of responses per prompt, number of classes $B$\\
   \STATE {\bfseries Output:} LLM for generating synthetic data $\Psi_{T + 1}$
   \STATE Embed private samples for each class $i=1...B$,  $E_i = \Gamma(\{s\}_{F(s) = i, s \in S})$ where $F(s)$ is the class index of sample s
   \STATE Initialize LLM $\Psi_1 = \Psi$

   \FOR{$t\leftarrow1$ $...$ $T$ }
      \STATE Initialize B response vectors $R=\{\emptyset, ... \emptyset\} = \{R^{(1)},...,R^{(B)}\}$
  \FOR {$b \leftarrow$ $...$ $B$}

   \FOR{$k\leftarrow 1 \ldots K$ }
   \STATE Generate prompt $\eta_k=\Lambda(\eta, b)$, 
   \STATE Generate $J$ responses  $R_{kj}^{(b)} = \Psi_t(\eta_k)$, $j\in [J]$
   \ENDFOR
   \STATE Calculate embeddings $E_{syn,t}^{(b)}=\{\Gamma(R_{kj}^{(b)})\}_{k\in [K], j
   \in [J]}$
   
   \STATE ${\rm Scores}_{t} \leftarrow$ ${\tt CENTRALSCORE}(E_{syn,t}^{(b)}, E_b) + \mathcal{N}(0, \sigma^2 {I})$
   \STATE Set $P[k,j]$ as the $j$-th highest score response for prompt $\eta_k$, according to  ${\rm Scores}_{t}$
   \STATE Initialize preference dataset $\mathcal{P}_t^{(b)} = \emptyset$
   \FOR{$k\leftarrow 1 \ldots K$ }
   \STATE Select positive synthetic sample: $\mathcal P_{t}^{(b)} [k,1] = P_t[k,1]$
   \STATE Select negative synthetic sample: $\mathcal P_{t}^{(b)} [k,2] = P_t[k,\ell]$
   \STATE Set prompt $\mathcal P_{t}^{(b)} [k,3] = \eta_k$
   \ENDFOR
   \ENDFOR
   \STATE Fine-tune: $\mathcal P_{t} = \bigcup_{b = 1}^B P_{t}^{(b)}$,  $\Psi_{t+1} \leftarrow \text{DPO}(\Psi_{t}, \mathcal P_{t})$ 
   \ENDFOR
   \STATE
   Output $\Psi_{T + 1}$

\end{algorithmic}
\end{algorithm}

Below is the similarity scoring function we use for the federated setting.

\begin{algorithm}[H]
\caption{{\tt SIMILARITY}}
   \footnotesize
   \label{alg:DPNN}
\begin{algorithmic}[1]
   \STATE {\bfseries Input:} Set of embeddings of private client data $E_i = \{emb(s_1^{(i)}), \ldots, emb(s_{m_i}^{(i)})\}$ for ${i\in \mathcal S^t}$, embeddings of synthetic data ${\rm E}_{syn}$, total synthetic samples $M=K\times J$ \\
    $\rm{Scores}$ $\leftarrow \boldsymbol 0^{M}$
   \STATE ${\rm Scores}[j]$ = $\frac{1 }{ m_i }\sum_{e_{pri} \in E_i} \frac{\langle e_{pri}, e_j \rangle}{\|e_{pri}\|\|e_j\|} $ for $e_j \in E_{syn}$ 
   \STATE ${\bf return}$ $
\mathbf{\text{Scores}}/{\max\!\bigl(1,\;\lVert\mathbf{\text{Scores}}\rVert_2\bigr)}$
\end{algorithmic}
\label{alg:dpnn}
\end{algorithm}

Below is the similarity scoring function we use for the central DP setting.

\begin{algorithm}[H]
\caption{{\tt CENTRALSCORE}}
   \footnotesize
   \label{alg:centraldpnn}
\begin{algorithmic}[1]
   \STATE {\bfseries Input:} Embeddings of private data $E$, embeddings of synthetic data ${E}_{syn}$ \\
    $\rm{Scores}$ $\leftarrow \boldsymbol 0^{M}$
   \STATE ${\rm Scores}[j]$ = $(1 / |E|) \sum_{e_{pri} \in E} \frac{\langle e_{pri}, e_j \rangle}{\|e_{pri}\|\|e_j\|} $ for $e_j \in E_{syn}$
   \STATE ${\bf return}$ Scores
\end{algorithmic}
\label{alg:dpnn-central}
\end{algorithm}

\section{Implementation Details of \name}
\label{sec:align-FL implementation}
\subsection{Model and Hyperparameters}
We choose LLaMA-3-8B as the data generator in \name and we fine-tune it iteratively during the course of the algorithm. To fine-tune the LLaMA-3-8B model, we use LoRA fine-tuning with rank 4, $\alpha$ = 8, applied to all the projection matrices in LLaMA-3-8B. We adapt the AdamW optimizer with a cosine learning rate scheduler with the learning rate ranging from $3\cdot10^{-7}$ to $8\cdot10^{-7}$. In the Congress and bioRxiv evaluations, the sample set $\Omega$ is a subset of the c4 dataset \citep{c4}, which is a large scale dataset from 2019, which we use for fair comparison with Private Evolution (PrE-Text), though we do not know their exact initial sample set because they did not release it. For the PubMed evaluation, the sample set $\Omega$ is a set of 2000 samples generated using the PubMed generation prompt in Table 16 of the Aug-PE paper, generated by LLaMA-3-8B-Instruct (which has a knowledge cutoff of March 2023), for comparison with Aug-PE \citep{xie2024differentially}. For each iteration, we fine-tune the models for 2 epochs and select the best checkpoint with the lowest FID score relative to the validation dataset. This checkpoint is used for synthetic data generation and as the starting point for the next iteration. The batch size is set to 24. 

In each round we generate 18000 synthetic data samples for the clients to evaluate. This is accomplished with 1800 prompts, each generating 10 samples for clients to rank. We select the 1st and 5th ranked sample for a given prompt for the ``selected" and ``rejected" data samples in the DPO preference dataset. We describe the experiments regarding which rank to use for constructing the preference dataset in detail in Appendix Section~\ref{sec:rejected_sample}.
To test the scaling relation with the number of clients per round and the total number of clients participating in the training, we set up the parameters and privacy budget shown in Table~\ref{table:setting}. The `all-MiniLM-L6-v2' sentence transformer model is used as the embedding model in \name. We note that we adopt ``sentence-t5-base” sentence transformer for PubMed during the step of fine-tuning ${\rm BERT}_{small}$, which follows the setting in AUG-PE. We ensure \name follows privacy guarantee of $(\epsilon, \delta)$-DP = (1, $3\times 10^{-6}$) or (7, $3\times 10^{-6}$) for both the bioRxiv and the Congressional Speeches datasets and run with 20 iterations for DP-FedAvg, DP-FTRL, PrE-Text for comparison. For AUG-PE, we set $(\epsilon, \delta)$-DP = (1, $2.72\times 10^{-6}$) or (4, $2.72\times 10^{-6}$). PubMed experiments are run with 10 iterations.

In terms of models for downstream tasks: 
\begin{itemize}
\item For BioRxiv $\&$ Congressional  Speeches, we fine-tuned the pre-trained DistillGPT2 for next-token prediction. We set the max sequence length as 64, number of generated synthetic data as 1,000,000, the batch size as 160, the learning rate as $2e^{-4}$, and the number of epochs as 80.
\item For PubMed, to compare with \citep{yue2023synthetictextgenerationdifferential}, we follow their procedure to leverage pre-trained ${\rm BERT}_{small}$ \cite{turc2019wellreadstudentslearnbetter}. We set the max sequence length as 512, number of generated synthetic data as 2000, batch size as 32, learning rate as 3e-4, the weight
decay as 0.01, and the number of epochs as 10. To compare with \cite{xie2024differentially}, we set up the $(\epsilon,\delta)$-DP value and hypterparameter according to their choice. For example, they set $\delta = {1 \over N_{priv}\cdot {\rm log}(N_{priv})}$ following \cite{yue2023synthetictextgenerationdifferential}. To achieve $\delta$ = $\{$1,4$\}$, we use noise multiplier $\sigma$ = $\{$13.7, 3.87$\}$ for 10 iterations under DP on all PubMed data. Note that our noise multiplier values are slightly different than \cite{xie2024differentially} due to different methods for calculating differential privacy. 
\end{itemize}

\subsection{Prompt Design}
\label{app:prompt}
\begin{figure*}[t]
\begin{center}
\centerline{\includegraphics[width=12.0cm]{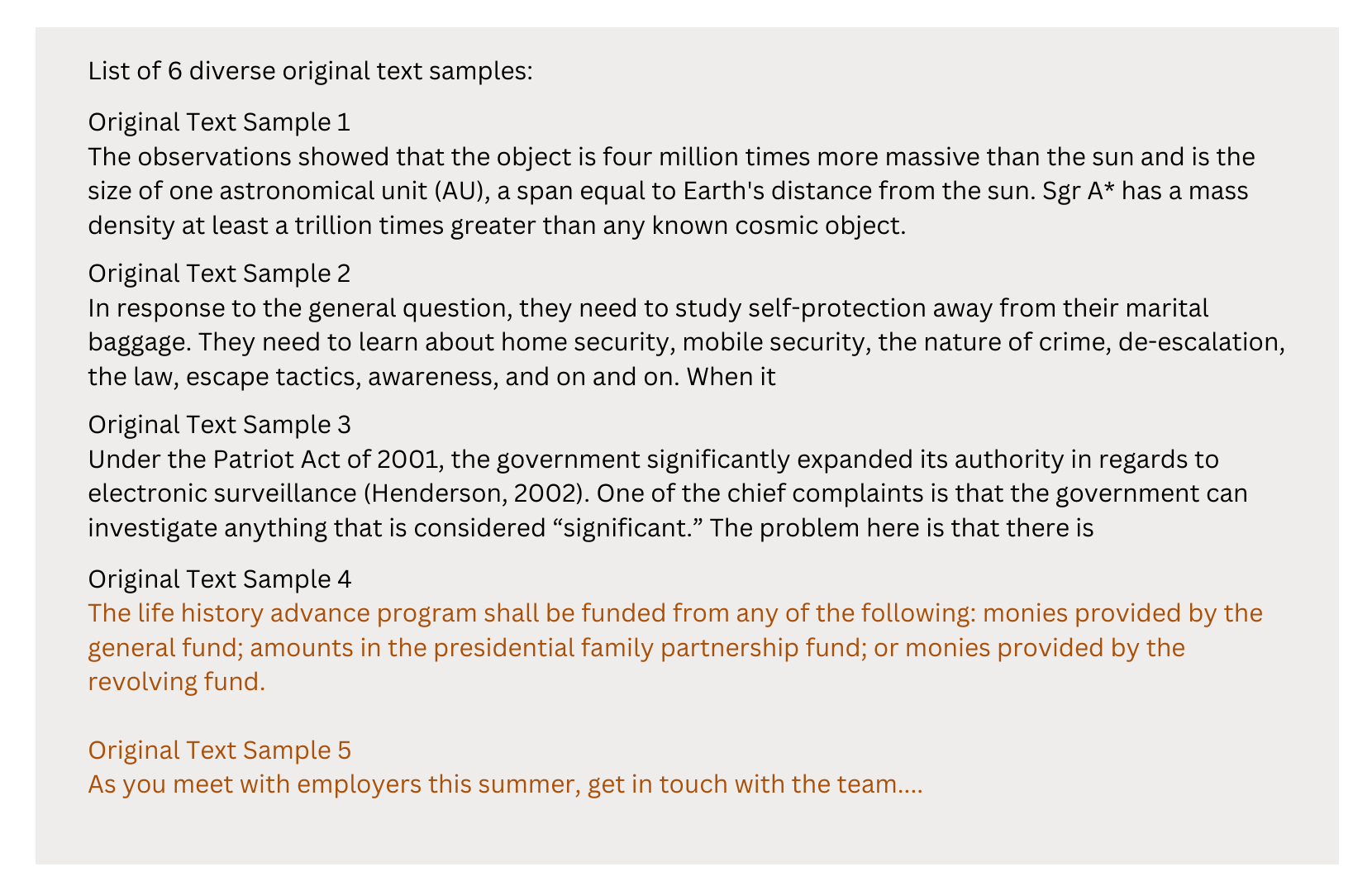}}
\caption{The synthetic data generation prompt for \name. The black text marks the input prompt, and the brown text after ``Original Text
Sample 4” is generated. The generated text between ``Original Text Sample 4" and ``Original Text Sample 5" is collected and used as a synthetic sample.}
\label{fig:prompt}
\end{center}
\vskip -0.2in
\end{figure*}

To compare with other data generator methods, we adopt the prompts used in the baseline models against which we compare. We generate the synthetic data using an approach similar to that in PrE-Text \cite{PrE-Text}. Figure~\ref{fig:prompt} shows an example of the prompt we use for prompting LLaMA-3-BB for generating synthetic data. For bioRxiv/Congress, we randomly take text samples from the c4 \citep{c4} dataset as our examples in the prompt. For PubMed, while running \name, we still adopt the prompt shown in Figure~\ref{fig:prompt} but reduce the number of examples to two in order to accommodate longer sequence lengths, randomly sampling generated abstracts from LLaMA-3-8B. For OpenReview, we prompt the model directly to generate paper reviews (similarly to \citep{xie2024differentially}).

\section{Implementation Details of Baseline Models}
\label{sec:baseline implementation}
In this section we provide implementation details for the baseline algorithms. We use two DP-FL baselines: DP-FedAvg and DP-FTRL. For the PE baseline, we implement PrE-Text \cite{PrE-Text} for the evaluations on the bioRxiv and Congressional Speeches datasets. For the PE baselines on the PubMed dataset we directly compare against the Aug-PE results from \citet{xie2024differentially}.

\subsection{DP-FedAvg}
We employ the FedAvg federated optimization algorithm \cite{mcmahan2017learning} to fully fine-tune DistilGPT2, avoiding linear probing due to its poor performance in DP language models \cite{lin2021privacy}. Our training configuration includes a batch size of 2, a sequence length of 64, and 20 rounds for \cref{tb:result} and 50 rounds for \cref{fig:scaling-eps1}, and either full or partial client participation. For differential privacy (DP), we utilize secure aggregation \cite{secagg} and introduce Gaussian noise \cite{mcmahan2017learning}. We evaluate the model using next-token prediction accuracy across various numbers of training epochs on the clients. We tune the learning rate within the range [0.001, 0.1, 0.1] and the clipping threshold between [0.01, 0.1, 1.0], selecting the model with the best performance on the evaluation set for reporting. The noise is scaled to ensure a privacy guarantee of $(\epsilon, \delta)$-DP where $\delta$ = 3$\cdot10^{-6}$ and $\epsilon$ = $\{$1,7$\}$, representing two distinct privacy regimes. The noise multipliers are $\sigma$ = $\{$19.3, 3.35$\}$ when considering all the data, and the settings for partial participation experiments are shown in Table~\ref{table:setting}.
\subsection{DP-FTRL}
We also use the DP variant of Follow-The-Regularized-Leader (DP-FTRL) algorithm \cite{kairouz2021practical} to fully fine-tune DistilGPT2. The hyperparameter settings are similar to DP-FedAvg other than the noise multipliers. The noise multipliers are $\sigma$ = $\{$19.5, 3.35$\}$ when considering all the data, and the settings for partial participation experiments are shown in Table~\ref{table:setting}.
\subsection{PrE-Text}\label{app:PrE-Text}
We follow similar settings as \citet{PrE-Text} with some modifications. The privacy budget is similar to DP-FedAvg and \name, with a privacy guarantee of $(\epsilon, \delta)$-DP where $\delta$ = 3$\cdot10^{-6}$ and $\epsilon$ = $\{$1,7$\}$ with $\sigma$ = $\{$19.3, 3.35$\}$ for full participation and partial participation in Table~\ref{table:setting}. We set the thresholds H = 0.1626, T = 20, and $N_{syn}$ = 1024. We adopt the ``all-MiniLM-L6-v2" sentence transformer model for text embedding generation.

\section{Experimental Details}
\subsection{Privacy Accounting}
\label{app:privacy-details}
\begin{table*}[t]
\scriptsize
\caption{Experiment privacy budget settings.}
\label{table:setting}
\vskip 0.15in
\begin{center}
\begin{NiceTabular}{ccccccc}
\toprule
Total $\#$ of & $\#$ of clients &  \multirow{2}{*}{$\sigma_{1}$\tabularnote{For DP-FedAvg, PrE-Text, \name.}, $\epsilon$ = 7}  & \multirow{2}{*}{$\sigma_{1}$\tabularnote{For DP-FedAvg, PrE-Text, \name.}, $\epsilon$ = 1}   &\multirow{2}{*}{$\sigma_{2}$\tabularnote{For DP-FTRL}, $\epsilon$ = 7} & \multirow{2}{*}{$\sigma_{2}$\tabularnote{For DP-FTRL}, $\epsilon$ = 1}  \\
clients & per round &  & &  & \\
\midrule
10000 & 1000 & --- & 3.4 & --- & 19.5\\
10000 & 5000 & --- & 15.5 & --- & 30.8\\
10000 & 10000  & --- & 30.6 & --- & 30.8 \\
\midrule
72000 & 72000 & 3.35 & 19.3 &3.35 & 19.5\\
133000 & 133000 & 3.35 & 19.3 &3.35 & 19.5\\

\bottomrule
\end{NiceTabular}
\end{center}

\vskip -0.1in
\end{table*}

The precise privacy settings we use and their corresponding $\epsilon$ values, as calculated by their corresponding privacy budget computation methods, 
are reported in \cref{table:setting}.
DP-FedAvg \citep{mcmahan2017learning} and Private Evolution (PrE-Text) \citep{PrE-Text} both use the Gaussian mechanism, and thus use similar computations. In both cases, we use the privacy accountant of the Opacus library \cite{yousefpour2021opacus}.
For DP-FedAvg, we calculate privacy by inputting the number of rounds, the client sampling ratio, setting the noise multiplier to be the product of $\sigma$ and the clipping threshold, choosing a $\delta \ll 1/|\mathcal S|$, and setting $\sigma$ for the desired $\epsilon$. Private Evolution (PrE-Text) \citep{PrE-Text} also uses the Gaussian mechanism, so we use the same accounting except the noise multiplier is the product of $\sigma$ and the maximum number of samples per client. 
For DP-FTRL, we follow the privacy accounting methods from their implementation. For Private Evolution (Aug-PE) \citep{xie2024differentially}, we report their reported $\epsilon$ directly.

\subsection{Evaluation Details for Different Datasets}
\label{app:dataset-eval}

\subsubsection{\benchname Evaluation}
For the bioRxiv and Congressional Speeches datasets, we use the PrE-Text version of Private Evolution because the PrE-Text evaluation focused on datasets with samples with max sequence length of 64. 

\subsubsection{PubMed and OpenReview Evaluation}
\label{app:pubmed}
For PubMed and OpenReview, our Private Evolution baseline  compares to Aug-PE, which has already been evaluated on PubMed and OpenReview \cite{xie2024differentially}. 
Note that PubMed and OpenReview was used by \citet{xie2024differentially} to evaluate central DP algorithms. In the central DP setting, there are no clients; all private data is held at the server and the goal is to release a model with DP guarantees. The notion of neighboring dataset in central DP is a centrally held dataset that is the same except for a single data sample. To compare our algorithm directly with results reported for Private Evolution (Aug-PE) \citep{xie2024differentially}, we replicate the central DP setting for this dataset by having one PubMed abstract per client and sampling all clients every iteration (or ``round'', in our case).

\section{Ablation Studies}
\subsection{Cosine similarity vs Nearest neighbors histogram}
In this section we perform an ablation justifying the choice of cosine similarity as a scoring function over the nearest neighbor histogram employed by Private Evolution. We find that using cosine similarity works much better than nearest neighbors histogram for our use case, because nearest neighbors histogram is too sparse to ensure the construction of meaningful preference pairs for \name.
\begin{figure*}[ht]
\begin{center}
\centerline{\includegraphics[width=14.0cm]{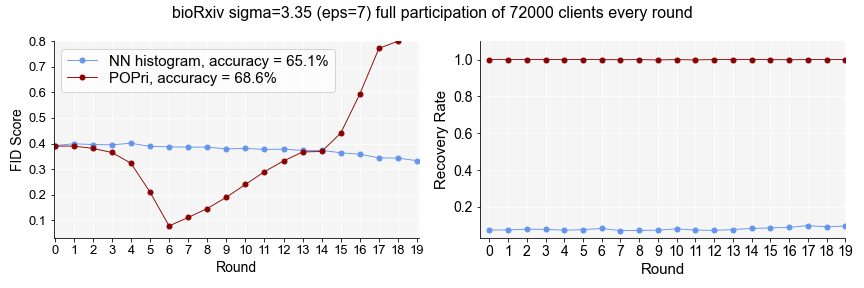}}
\caption{\textbf{Left:} FID scores of POPri using NN histogram scoring vs. POPri using cosine similarity. \textbf{Right:} After the client feedback stage, we measure the percentage of the time the non-noised and non-clipped score (nearest neighbor histogram scoring or cosine similarity scoring) of the chosen sample is higher than the rejected sample. For cosine similarity, this ``recovery rate'' is much higher (nearly 100\%) than in nearest neighbors histogram. \textbf{Interpretation.} Nearest neighbors histogram is much sparser than cosine similarity, often assigning zero to all synthetic samples associated with a given prompt in POPri. This leads to preference pairs often being completely noisy. Cosine similarity provides denser scoring that allows the construction of meaningful preference pairs for all prompts.
}
\label{fig:cosine_vs_nnhist}
\end{center}
\vskip -0.2in
\end{figure*}

\begin{figure*}[ht]
\begin{center}
\centerline{\includegraphics[width=8.0cm]{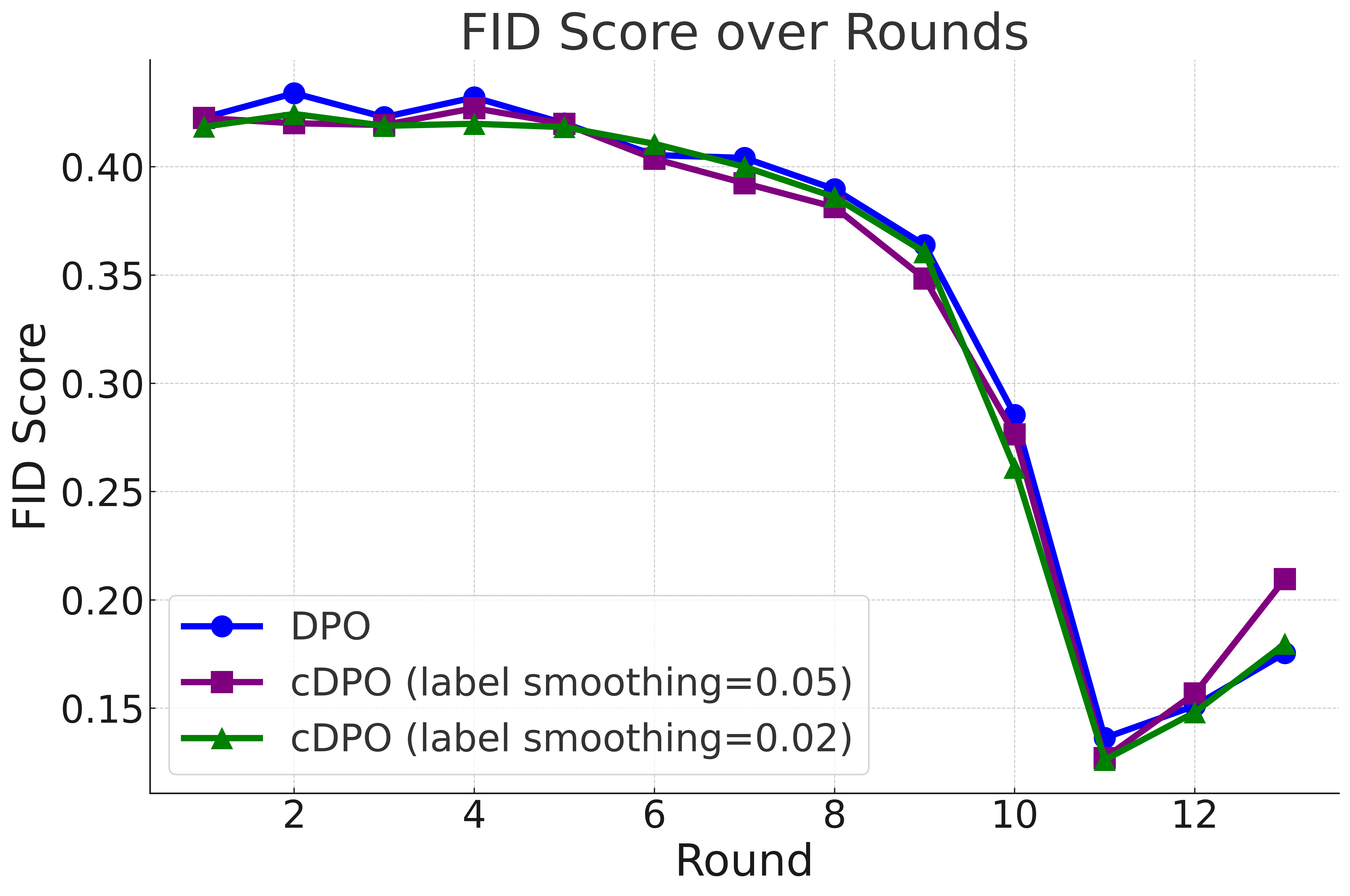}}
\caption{In this experiment, we investigate whether the use of label noise-resistant alignment methods could allow the use of higher-ranked rejected samples. To do this, we used the third-ranked sample as the rejected sample, and evaluated different settings for conservative DPO (cDPO) \citep{mitchell2023cdpo}. We used the bioRxiv dataset experiment setting, set eps=7, learning rate = 8e-7. We find that by tuning the level of conservative-ness we may be able to improve slightly on vanilla DPO.
}
\label{fig:cdpo}
\end{center}
\vskip -0.2in
\end{figure*}
\subsection{Alignment methods}
\label{sec:ipo}

We also experiment with a noise-resistant (or robust) DPO method, conservative DPO (cDPO) \citep{mitchell2023cdpo}, to see if by using it we can select a higher ranked rejected sample (recall we use the 5th ranked, and higher ranked samples would introduce more noise into the preference pairs). In \cref{fig:cdpo} we find that it can help slightly when choosing a higher rejected sample ranking.

\subsection{Rejected sample selection}
\label{sec:rejected_sample}
\begin{figure}[ht]
\begin{center}
\centerline{\includegraphics[width=14.0cm]{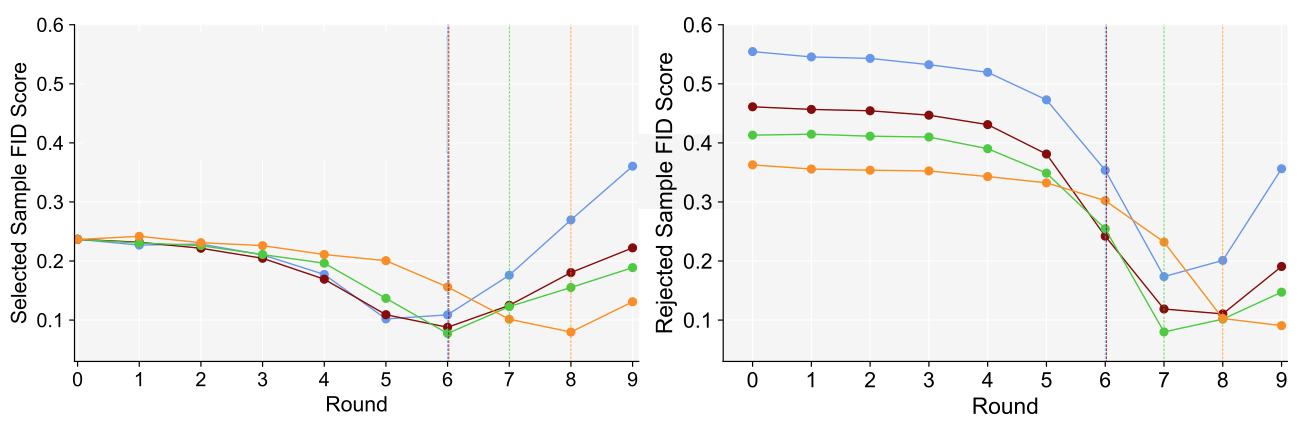}}
\caption{Ablation study for selecting rejected samples in the preference data. Here we generate 10 samples for each prompt and select Nth ranked data as the rejected sample, where N is 5, 7, or 10. The vertical lines indicate the round at which the best next-word-prediction accuracy was achieved for each choice of rank. Note that the model that produces the lowest overall FID (not the lowest selected sample FID or the lowest rejected sample FID) is the best synthetic data generation model, since on the final round all generated samples are utilized to form the synthetic dataset. We hypothesize that round 7 corresponds to the highest accuracy for the rank 5 model because after that point, the selected sample FID is higher than the rejected sample FID, which would mean the preference dataset has become mis-aligned with the objective of generating good synthetic data.}
\label{fig:rank}
\end{center}
\end{figure}

We construct the DPO preference data via client feedback by generating ten samples from the same prompt and then picking the ``selected" and the ``rejected" samples. The samples with the highest scores among the ten examples are picked as the ``selected" sample in the DPO preference dataset. We experiment on which rank should be utilized as the ``rejected" sample in the DPO preference dataset. In Fig~\ref{fig:rank} we further explore the effects by examining the ``rejected" and ``selected" sample FID scores as a function of round. In the left panel where the ``selected" sample FID values are shown, their magnitude and trends behave similarly before they reach the best results (marked by colored dashed vertical lines). For the ``rejected" sample FID shown in the right panel, the 5th rank ``rejected" samples yield the lowest FID score and therefore smaller gap between the preference sample pairs. However, we also find that higher rank does not always yield better results. This may result from the boundary between the ``rejected" and ``selected" samples becoming undistinguishable for rank $<$ 5th due to DP noise. We therefore select 5th rank samples as our ``rejected" DPO preference samples.

\begin{figure*}[ht]
\begin{center}
\centerline{\includegraphics[width=13.0cm]{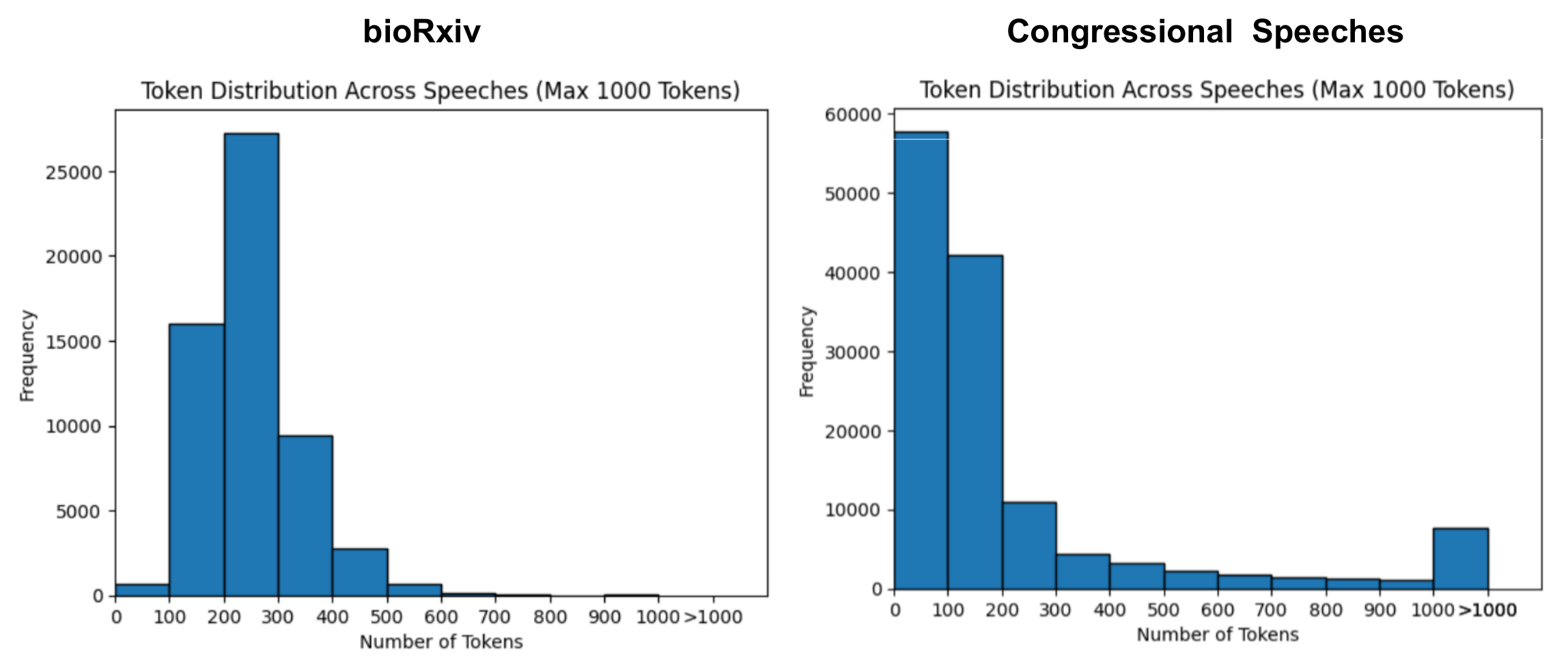}}
\caption{The distribution of how many tokens are in each client's dataset for the bioRxiv and Congressional Speeches datasets.}
\label{fig:dataset_token_distribution}
\end{center}

\end{figure*}
\section{Datasets}
\label{app:dataset}
\begin{table*}[t]
\scriptsize
\caption{Dataset details.}
\label{data-table}
\vskip 0.15in
\begin{center}
\begin{tabular}{lccccc}
\toprule
Dataset & $\#$ Train Samples& $\#$ Validation Samples & $\#$ Test Samples & Max Sequence Length & Average $\#$ of samples per client  \\
\midrule
bioRxiv & 72000  & 2000 & 1584 & 64 & 6.6 $\pm$ 2.6\\
Congressional Speeches & 133000 &4200 & 1547 & 64 & 5.0 $\pm$ 16.3\\
PubMed  & 75316 & 14423 & 4453 & 512 & 1\\
\bottomrule
\end{tabular}
\end{center}
\vskip -0.1in
\end{table*}

\begin{figure*}[h]
\begin{center}
\centerline{\includegraphics[width=5.0cm]{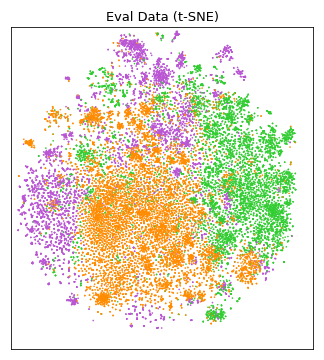}}
\caption{A t-SNE clustering of the Congressional Speeches dataset. US data is colored in purple, UK data is colored in orange, and Canada data is colored in green. We find that the three datasets form distinct clusters and also distinct sub-clusters.}
\label{fig:tsne_congressional}
\end{center}

\end{figure*}
\paragraph{bioRxiv.} This dataset consists of abstracts from bioRxiv papers with appropriate copyright permission from April 2023 to August 2024. This was done by using the bioRxiv public API to retrieve the abstracts of the paper with permitted licenses (i.e. `CC BY NC ND', `CC BY ND', `CC BY NC', `CC BY', `CC0'). This dataset consists of 72k abstracts (clients), each of which we split into chunks of 64 tokens to form samples.

\paragraph{Congressional Speeches.} This dataset consists of speeches from US, UK and Canada congressional/parliamentary transcripts from April 2023 to August 2024. 
All speeches are published under a permissive license which allows for third-party use (as detailed in the dataset cards). 
There are 134k speeches (clients) in total, and 1930 unique speakers. We collected this dataset by using public APIs to retrieve data from each country’s official congressional/parliamentary library website. Then we sanitized the data by removing (1) boilerplate procedural language, (2) sentences with more than 30\% of the characters not being letters, and (3) some written notation that does not correspond to spoken words. We split each speech into chunks of 64 tokens each. We believe that this dataset is a major contribution because spoken language may be more resistant to contamination (especially for the UK and Canada parliamentary debates). Because they are more conversational and have a large degree of improvisation (many debates are off-the-cuff), they are less likely to be generated by LLMs. Because Congressional Speeches contains a diverse collection of speeches across speakers and also countries, the dataset forms many distinct clusters, reflecting the diversity of the dataset (\cref{fig:tsne_congressional}).

We will update the dataset periodically with the latest data to allow future researchers to test their algorithms or ideas against an uncontaminated dataset.

\end{document}